%% file: main.tex
\documentclass[conference]{IEEEtran}
\usepackage{microtype} %runts
\usepackage{cite}
\usepackage{amsmath,amssymb,amsfonts}
\usepackage{algorithmic}
\usepackage{graphicx}
\usepackage{textcomp}
\usepackage{xcolor}
\input{math_commands.tex}

\input{other_commands.tex}
\usepackage{float} % For exact table placement with H
\usepackage{booktabs}
\usepackage{adjustbox} % Optional: For scaling and positioning control
\usepackage{hyperref}
\usepackage{url}
\usepackage[most]{tcolorbox}
\usepackage{graphicx}

\usepackage{mdframed}
\newmdenv[backgroundcolor=white, linecolor=black]{myframed}
\newtcolorbox[auto counter]{Insight}[1][]{title={\bfseries Insight~\thetcbcounter},enhanced,drop shadow={black!50!white},
  coltitle=black,
  top=0.1in,
  attach boxed title to top left=
  {xshift=1.5em,yshift=-\tcboxedtitleheight/2},
  boxed title style={size=small,colback=pink},#1}

\def\BibTeX{{\rm B\kern-.05em{\sc i\kern-.025em b}\kern-.08em
    T\kern-.1667em\lower.7ex\hbox{E}\kern-.125emX}}
\begin{document}

\title{RobustBlack: Challenging Black-Box Adversarial Attacks on State-of-the-Art Defenses\\
}

 \author{\IEEEauthorblockN{1\textsuperscript{st} Mohamed Djilani}
 \IEEEauthorblockA{\textit{SnT} \\
 \textit{University of Luxembourg}\\
 Luxembourg \\
 mohamed.djilani@uni.lu\\}
 \and
 \IEEEauthorblockN{2\textsuperscript{nd} Salah Ghamizi$^{\dagger}$}
 \IEEEauthorblockA{\textit{Luxembourg Institute of Health} \\
 \textit{SnT / University of Luxembourg}\\
 Luxembourg \\
 salah.ghamizi@lih.lu\\}
  
 \and
 \IEEEauthorblockN{3\textsuperscript{rd} Maxime Cordy}
 \IEEEauthorblockA{\textit{SnT} \\
 \textit{University of Luxembourg}\\
 Luxembourg \\
 maxime.cordy@uni.lu\\}
 
}

\maketitle

\begingroup
\renewcommand\thefootnote{$\dagger$}
\footnotetext{This work was conducted while at Luxembourg Institute of Health.}
\endgroup

\begin{abstract}

\input{pages/0-abstract}
\end{abstract}

\IEEEpeerreviewmaketitle

\input{pages/1-introduction}
\input{pages/2-related_works}
\input{pages/3.RQ1-V3}
\input{pages/4.RQ2-V3} 
\input{pages/5.RQ3-V3}

\input{pages/discussion}

\input{pages/6-conlusion}

\section*{LLM usage considerations}

LLMs were used for editorial purposes in this manuscript, and all outputs were inspected by the authors to ensure accuracy and originality. Similarly, LLMs were used to assist in the process of writing code, and searching for relevant publications and literature reviews. All code and publications were manually verified and tested to ensure coherence in the results.

\section*{Acknowledgements}
This research was funded in whole, or in part, by the Luxembourg National Research Fund (FNR), grant reference 18886361.
Dr. Ghamizi is supported by the Luxembourg National Research Fund (FNR) CORE C24/IS/18942843.

\bibliographystyle{IEEEtran}
\bibliography{bib/adv,bib/blackbox,bib/lgv,bib/other,bib/diffattack,bib/egbib,bib/rebuttal,bib/satml,bib/blackboxbench,bib/rebuttal_SATML}

\clearpage
\appendix
\input{pages/appendix}

% \clearpage
% \input{pages/cover_letter}

% \vspace{12pt}
% \color{red}
% IEEE conference templates contain guidance text for composing and formatting conference papers. Please ensure that all template text is removed from your conference paper prior to submission to the conference. Failure to remove the template text from your paper may result in your paper not being published.

\end{document}

%% file: math_commands.tex
\usepackage{amsmath,amsfonts,bm}

\def\eqref#1{equation~\ref{#1}}
\def\1{\bm{1}}

\DeclareMathAlphabet{\mathsfit}{\encodingdefault}{\sfdefault}{m}{sl}
\SetMathAlphabet{\mathsfit}{bold}{\encodingdefault}{\sfdefault}{bx}{n}

\newcommand{\R}{\mathbb{R}}

\def\argmaxop{\mathop{\rm arg\,max}\limits}

\def\R{\mathbb{R}}

%% file: other_commands.tex
\newcommand{\newtext}{\textcolor{black}} %blue

\newcommand{\reformulatedtext}{\textcolor{black}} %orange
\newcommand{\correct}{\textcolor{black}}  %orange

%% file: pages/0-abstract.tex
Although adversarial robustness has been extensively studied in white-box settings, recent advances in black-box attacks (including transfer- and query-based approaches) are primarily benchmarked against weak defenses, leaving a significant gap in the evaluation of their effectiveness against more recent and moderate robust models (e.g., those featured in the Robustbench leaderboard).
In this work, we argue that this gap is problematic and the previous benchmarks conclusions do not hold under more robust evaluation frameworks, leading to contradicting conclusions on the transferability of adversarial attacks.
We extensively evaluate the effectiveness of 13 popular black-box attacks, representative of the top ten popular transferability theories. We benchmark these attacks against eight top-performing and standard defense mechanisms on the ImageNet dataset. Our empirical evaluation reveals the following key findings: (1) the most advanced black-box attacks struggle to succeed even against simple adversarially trained models; (2) robust models that are optimized to withstand strong white-box attacks, such as AutoAttack, also exhibit enhanced resilience against black-box attacks; and (3) robustness alignment between the surrogate models and the target model can significantly impact the success rate of transfer-based attacks.

%% file: pages/1-introduction.tex
\section{Introduction}

\begin{table*}[ht]
    \centering
    \caption{Theories about the transferability of adversarial examples. \textbf{In bold when the evaluation protocol of the black-box is sufficiently robust}. We report the largest image dataset, the smallest perturbation distance, and the deep learning model families used in the evaluation. FNN for feed-forward neural network, CNN for convolutional neural network}
    
    \begin{tabular}{llcccc}
        \textbf{Theory} & \textbf{Dataset} & \textbf{Representative attacks} & \textbf{Attack budgets} & \textbf{Target architectures} & \textbf{Target defenses} \\

       \midrule
       Linearity \cite{goodfellow2014explaining} & \textbf{ImageNet} & FGSM & $L_2=0.25$ & Convolution & FGSM AT \\
       Decision Boundary Similarity \cite{liu2016delving} & \textbf{ImageNet} & FGSM & $L_0=23$ & Convolution & $\varnothing$ \\
       Gradient Similarity \cite{demontis2019adversarial} & MNIST & FGSM & $L_2=1$ & FNN & $\varnothing$ \\
       Adversarial Subspace \cite{tramer2017space}& MNIST & FGSM & $L_2=4$ & FNN, CNN & FGSM AT \\
       Non-Robust Features \cite{ilyas2019adversarial} & \textbf{ImageNet} & PGD & $L_2=3$ & CNN & PGD AT \\
       Model complexity \cite{wu2020towards} & \textbf{ImageNet} & IFGSM & $L_2=15$ & CNN & $\varnothing$ \\
       Perturbation interactions \cite{wang2020unified} & \textbf{ImageNet} & PGD & $L_{\infty}=16/255$  & CNN & FGSM AT \\
       Spectral properties \cite{cheng2021spade} & CIFAR-10  & PGD & $\mathbf{L_{\infty}= 2/255}$ & FNN, CNN & PGD AT \\
       Knowledge transferability \cite{liang2021uncovering} &  STL-10 & PGD & $\mathbf{L_2= 2/255}$ & CNN & $\varnothing$ \\
       Generalization gap \cite{wang2023role} & CIFAR-100  & PGD & $L_{\infty}=8/255$ & CNN & $\varnothing$ \\
       %Bound theory \cite{fan2024transferability} & \textbf{ImageNet} & \textbf{TPA} & $\mathbf{L_{\infty}=4/255}$ & CNN & PGD AT + Data augmentation \\
       \midrule
       \textbf{Proposed protocol}  & \textbf{ImageNet} & \textbf{AutoAttack, BASES} & $\mathbf{L_{\infty}=4/255}$ & \textbf{CNN, Transformers} & \textbf{ Robustbench}  \\

    \end{tabular}
    \label{tab:theories}
\end{table*}

Since the discovery that deep learning models are susceptible to minor input disturbances, resulting in adversarial examples (AE) \cite{goodfellow2014explaining}, the development of robust models has become one of the
most active topics in the machine learning community. This topic has been thoroughly explored for white-box settings, settings where attackers have full knowledge of the target, leading to powerful standardized attacks \cite{croce2020reliable} and benchmarks \cite{croce2021robustbench}.

In parallel, more "realistic" studies of adversarial vulnerability involved restricting the knowledge or the capabilities of the attacker. These scenarios are referred to as black-box scenarios (or gray-box when partial knowledge of the target is available \cite{guo2018countering}). 
State-of-the-art (SoTA) black-box attacks leverage one or both of the following mechanisms: (1) adversarial examples transferability and (2) iterative queries with meta-heuristics. 

Transferability methods craft adversarial examples on a specifically built surrogate model (with white-box access) such that this example should transfer to (i.e. also fools) a target model (to which they have black-box access). Iterative query methods use query feedback from the target model to guide or refine the optimization process to uncover its vulnerabilities.

Black-box attacks are powerful tools to study real-world attack scenarios and to better understand the mechanisms behind adversarial vulnerabilities of deep learning models, in particular through the study of attacks' transferability. 
However, they still rely on outdated and unstandardized evaluation practices \reformulatedtext{(cf. Table \ref{tab:theories})}. In contrast to white-box research, recent black-box attacks are still designed and evaluated using large epsilon noise budgets, small architectures, and non-robust target models.

Traditional theories such as the decision boundary similarity \cite{liu2016delving}, the adversarial subspace theory \cite{tramer2017space}, and the non-robust features \cite{ilyas2019adversarial} are still well recognized, but have been demonstrated using simple transfer attacks, with very large budgets and non-robust models.
%Even the recent work by Fan et al. \cite{NEURIPS2024_flatness_transfer}, which challenged the conventional theory linking adversarial flatness to transferability (using a new bound theory), relied in their evaluation on $\epsilon=16/255$ and simple adversarial training mechanisms. 
In contrast, Robustbench~\cite{croce2021robustbench} recommended much stronger models and enforced a budget of $\epsilon=4/255$ as a reasonable evaluation budget.
 
 \textbf{At least ten} of the main theories have been empirically demonstrated using incomplete black-box attacks protocols (Table \ref{tab:theories}): linearity \cite{goodfellow2014explaining}, gradient similarity \cite{demontis2019adversarial}, adversarial subspace \cite{tramer2017space}, non-robust features \cite{ilyas2019adversarial}, model complexity \cite{wu2020towards}, adversarial perturbation interactions \cite{wang2020unified}, spectral properties \cite{cheng2021spade}, knowledge transferability \cite{liang2021uncovering}, generalization gap \cite{wang2023role}, and decision boundary similarity \cite{liu2016delving,hwang2024similarity}.%, and bound theory \cite{NEURIPS2024_flatness_transfer}.

Our work does not intend to challenge all previous work and theories about the vulnerability of deep learning models to adversarial attacks; indeed, many theories were evaluated with the best protocols at that time. We would rather like to demonstrate that adhering to more stringent evaluation protocols leads to different insights and conclusions on the vulnerability of the models and effectiveness of the black-box attacks. 

In addition, given the importance of black-box attacks in assessing real-world robustness of ML systems put in production, we also hypothesize that the previous evaluation results of black-box attacks can be misleading. Even old robustification mechanisms enable us to successfully evade supposedly SoTA black-box attacks. LGV \cite{gubri2022lgvboostingadversarialexample}, one of the strongest transfer attacks, sees its success rate drop from $87.39\%$ to $2.17\%$ under a 2020 defense by Salman et al.\cite{salman2020adversarially}. \reformulatedtext{Our first insight suggests to \textbf{reconsider} previous claims of effectiveness of black-box attacks and demonstrates that simple adversarial training is sufficient against the majority of SoTA black-box attacks.}

\reformulatedtext{While our first insight challenges the effectiveness of SoTA attacks, we tackle the benefits of SoTA defenses in our second study. We evaluated 9 defenses from Robustbench white-box ImageNet benchmark \cite{robustbench} and categorized these defenses into four families. Our analysis demonstrates that within the same family of defenses, architectures and model size have limited impact on the effectiveness of defenses. Our main insight from this study is that SoTA defenses are starting to plateau and show \textbf{diminishing returns} against SoTA black-box attacks. }  

Finally, our work argues that an incomplete evaluation of the interactions between new adversarial defenses and black-box attacks can actually lead new defenses to strengthen already existing black-box attacks, opening new surface attacks that were not considered before. For example, BASES \cite{bases} can leverage recent robust models from Liu et al.\cite{liu2024comprehensive} to increase its effectiveness against the most robust models from $1.26\%$ to $12.55\%$. \reformulatedtext{In fact, we analyzed the top 20 defenses in the  Robustbench ImageNet leaderboard ($L_{\infty}$, eps=4/255), and none have explored how their design choices might inadvertently aid black-box attacks}. Even Blackboxbench, the most recent and complete benchmark of black-box attacks \cite{zheng2025blackboxbench}, did not consider the \textbf{interactions} between SoTA white-box defenses and SoTA black-box attacks. \reformulatedtext{Our last insight highlights the \textbf{new risks} posed by SoTA defenses, that can be leveraged to improve existing (weak) attacks.}

To the best of our knowledge, this paper is the first to study the effectiveness of black-box attacks against standardized protocols and to demonstrate the need to confront black-box attacks to robust defenses and up-to-date evaluation settings. \correct{ Our contributions can be summarized as follows:}
\begin{enumerate}

\item We demonstrate that simple adversarial training mechanisms reduce the effectiveness of black-box attacks proven effective against standard models, underscoring the need for more advanced black-box attack strategies to address the robustness of real-world models and systems.

\item We show that white-box robustness could serve as a proxy for black-box robustness. Defenses optimized against AutoAttack generalize well to black-box scenarios. 

\item We demonstrate that these effective defense mechanisms can inadvertently contribute to enhancing the success rate of black-box attacks. By using robust models as surrogates, attacks can generate adversarial examples more likely to transfer to robust models, leading to an average increase in success rate of 6.49 percentage points across attacks and target models.

\end{enumerate}

Our contributions send a clear message to the adversarial research community on the importance of considering white-box defenses when designing or evaluating black-box attacks, or exploring transferability mechanisms and theories. They also urge researchers working on defense mechanisms to consider how their recent defenses can help malicious parties improve existing black-box attacks, and study how that their work could lead to stronger attacks against seemingly robust defenses. 
We provide a replication package with our experiments and benchmark on \url{https://figshare.com/projects/RobustBlack/265786}.

%% file: pages/2-related_works.tex
\section{Background}

\subsection{Preliminaries}
\label{subsec:prelim}
\paragraph{Adversarial perturbation~\cite{croce2021robustbench}} Let $x \in \R^d$ be an input point and $y \in \{1, \ldots, C\}$ be its correct label. For a classifier $f: \R^d \rightarrow \R^C$, we define a \textit{successful adversarial perturbation} with respect to the perturbation set $\Delta \subseteq \R^d$ as a vector $\vec{\delta} \in \R^d$ such that
    $$\argmaxop_{c \in \{1, \ldots, C\}} f({x}+\delta)_c \neq y\quad \textrm{and} \quad {\delta} \in \Delta,$$
where the perturbation set $\Delta$ is chosen such that all examples in $x+\delta$ have $y$ as their true label.
This motivates a robustness measure called \textit{robust accuracy}, which is the fraction of data points on which the classifier $f$ predicts the correct class for all possible perturbations from the set $\Delta$.
Computing the exact Robust Accuracy (RA) is in general intractable and, when considering $\ell_p$-balls as $\Delta$, NP-hard even for single-layer neural networks. 
In practice, an \textit{upper bound} for robust accuracy is determined by \textit{adversarial attacks}, generally involving optimization of a differentiable loss function or a reward through search algorithms that aim to identify a successful adversarial perturbation. The tightness of the upper bound depends on the strength of the attack.
\paragraph{Success rate} Given that different models show varying clean performances (e.g., test accuracy on the original set), robust accuracy will be impacted by the initial performance as much as the intrinsic robustness of the models. Thus, we base our study on an agnostic test performance metric: attack success rate (ASR).
We define ASR for a classifier $f$ under attack $\mathcal{A} ; \mathcal{A}(x) = {x}+\delta$ as

$$
\text{ASR}(f,\mathcal{A}) = \mathbb{E}_{(x,y)\sim\mathcal{D}}\left[\mathbb{I}\left(f(\mathcal{A}(x)) \neq f(x)\right)\right].
$$

For a model with a 100\% test accuracy, the two metrics are related $ \text{ASR} = 1 - \text{RA} $.

%\paragraph{Perturbation budget}  }

%\paragraph{Attack budget} XXX}

\subsection{Black-Box Attacks: Transfer-Based and Query-Based Methods}
\label{subsec:black-box}
Black-box attacks have seen notable advances, especially within transfer-based and query-based approaches. Transfer-based attacks rely on adversarial examples generated by a surrogate model, which are then tested on a target model. Early approaches like Projected Gradient Descent (PGD) \cite{madry2017towards} laid the groundwork for such attacks by iteratively optimizing perturbations to maximize the success rate on the target model. This foundational method was enhanced by the Momentum Iterative Fast Gradient Sign Method (MI-FGSM), which incorporated momentum into PGD to improve transferability and stability across iterations \cite{dong2018boosting}. Further innovations, such as the Diverse Input Fast Gradient Sign Method (DI-FGSM) and Translation-Invariant FGSM (TI-FGSM), introduced input transformations and image translations, respectively, to increase the robustness and generalizability of adversarial examples across different models \cite{xie2019improving, dong2019evading}.

Subsequent methods such as Variance Tuning (VMI, VNI) refined gradient calculations considering gradient variance between iterations \cite{wang2021enhancing}, leading to more effective adversarial perturbations. Similarly, ADMIX leveraged diverse inputs by mixing the target image with randomly sampled images to generate more transferable attacks \cite{wang2021admix}. The Skip Gradient Method (SGM) focused on model architecture, using skip connections which facilitate the creation of adversarial examples that have high transferability \cite{wu2020skip}. Universal Adversarial Perturbations (UAP) took a different approach by focusing on universality, generating a single small perturbation that can effectively disrupt a classifier in most natural images \cite{moosavi2017universal}. In particular, Ghost networks (GHOST) and Large Geometric Vicinity (LGV) contributed to transferability by modifying skip connections in surrogate models and by exploring variations in surrogate models' vicinity, respectively \cite{li2020learning, gubri2022lgvboostingadversarialexample}.

In query-based attacks, which generally have higher success rates than {transfer attacks~\cite{andriushchenko2019square}}, methods like Zeroth Order Optimization (ZOO) \cite{brendel2018decision,}, Decision Boundary Attack (DBA) \cite{ilyas2018black}, Natural Evolutionary  Strategies (NES) \cite{chen2020hopskipjump},  HopSkipJump \cite{chen2020hopskipjump}, \newtext{  SignFlip \cite{Chen2020boosting}, and Square \cite{ACFH2020square}} rely on iterative query feedback to optimize perturbations, often not utilizing surrogate models to boost efficiency .

{The following attacks then focused on reducing the number of queries. \cite{andriushchenko2019square} introduced a query-efficient black-box attack using randomized square-shaped updates at image boundaries, SimBA \cite{guo2019simba} proposed a simple attack using orthogonal search directions (e.g., DCT basis). Sign-OPT \cite{cheng2020signopt} introduces a hard-label attack estimating gradient signs instead of magnitudes, while SignHunt \cite{aldujaili2020signhunt} uses sign bits for gradient estimation. to minimize the number of queries. RayS \cite{chen2020rays} reformulates the boundary search as discrete optimization, eliminating gradient estimation.} 

{Recent work combined both surrogate representations and query feedback to create highly effective and efficient black-box attacks. Representative attacks include Transferable Model-based Embedding (TREMBA) and BASES \cite{huang2019black,cai2022blackbox}. BASES leverages an ensemble of surrogates with query feedback to dynamically adjust surrogate weighting, while TREMBA utilizes a generator trained with surrogate models, exploring latent space for more effective query attacks.}

\subsection{Defenses Against Strong Adversarial Attacks}
\label{subsec:defenses}

In parallel, adversarial defenses have evolved to mitigate the risks posed by increasingly sophisticated white-box attacks. As defenses became more robust, the need for a reliable, standardized benchmark to accurately assess their effectiveness grew increasingly apparent. In response, AutoAttack emerged as a reliable, parameter-free method to evaluate adversarial robustness, offering a computationally affordable and a standardized benchmark applicable to various models \cite{croce2020reliable}. Building on this, Robustbench established a leaderboard specifically to evaluate adversarial robustness, as outlined by \cite{croce2021robustbench}. This leaderboard includes results on ImageNet large-scale datasets \cite{deng2009imagenet}, and offers a ranking comparison of various defense strategies in specific perturbation budgets. Among the defenses featured on the leaderboard are those proposed by \cite{salman2020adversarially, singh2024revisiting, liu2024comprehensive, bai2024mixednuts, madry2017towards}, which offers a thorough assessment of the effectiveness of the leading defense mechanisms. Traditional defenses, such as Madry’s adversarial training, initially combined clean and adversarial data during training, establishing a baseline for robustness \cite{madry2017towards}. Recent advancements have incorporated complex pre-training schemes and data augmentations (e.g., ConvStem architectures, RandAugment, MixUp, and CutMix). \cite{singh2024revisiting} made alterations to the convnets architecture, particularly substituting PatchStem with ConvStem, and modified the training scheme to enhance the robustness against previously unencountered l1 and l2 threat models. \cite{liu2024comprehensive} further advances the training scheme with large-scale pre-training, combined with label smoothing, weight decay, and Exponential Moving Averaging (EMA) to enhance generalization against unseen adversarial examples. \cite{bai2024mixednuts} builds on {a given} defense framework \cite{liu2024comprehensive}, focusing specifically on balancing clean and robust accuracy through the nonlinear mixing of robust and standard models, addressing the often-seen tradeoff between clean and robust performance.

\subsection{Robustness Under Black-Box Setting}

% In this part, we focus on how our work introduces a novel contribution compared to previous studies on defenses against black-box adversarial attacks in image classification tasks. We begin by discussing five representative works for each, highlighting their key contributions, and then detail how our approach builds upon or complements these efforts, addressing gaps, and introducing new perspectives in this domain.

In this part, we focus on how our work introduces a novel contribution compared to previous studies on defenses against black-box adversarial attacks in image classification tasks. \reformulatedtext{We discuss the following representative works \cite{papernot2017practical, dong2020benchmarking, mahmood2021beware, mahmood2021robustness, ghaffari2022adversarial, zheng2025blackboxbench}}, highlighting their key contributions and how our approach builds on or complements these efforts by addressing gaps and introducing new perspectives in this domain.

\cite{papernot2017practical} introduced a black-box attack strategy targeting deep neural network (DNN) models and evaluated its effectiveness against adversarial training and defensive distillation. They varied the magnitude of the perturbation used during training and the attack phase and demonstrated that small perturbations of adversarial training in training led to gradient masking, which their attack could bypass by doubling the perturbation budget in the attacking phase, whereas larger perturbations in training improved robustness. In contrast, our work focuses on evaluating adversarially trained SOTA defenses against black-box attacks using a fixed small perturbation 4 over 255 in the attack phase.

\cite{dong2020benchmarking} evaluated adversarial robustness in image classification tasks against white-box and black-box attacks. Their work addressed various defense techniques such as robust training, input transformation, randomization, certified defenses, and model ensembling. Our study seeks to complement this by focusing specifically on SoTA robust training-based defenses. Moreover, our work explores a broader landscape of black-box attacks by including recent methods such as ADMIX, BASES, and TREMBA. In doing so, we provide an analysis of the leading robust training defenses against recent black-box adversarial strategies. 

\cite{mahmood2021beware} noted that the majority of existing defenses primarily address white-box attacks, neglecting the crucial aspect of black-box adversarial robustness. They provided a wide evaluation of adversarial defenses with a Convolutional Neural Network (CNN) architecture on benchmark datasets such as CIFAR-10 \cite{krizhevsky2009learning} and Fashion-MNIST \cite{xiao2017fashion} against 12 attacks. In a follow-up work, \cite{mahmood2021robustness} extended this investigation to Vision Transformers (ViTs), evaluating their adversarial robustness against white-box attacks and two black-box attacks on ImageNet. They suggested that ensemble defenses can enhance robustness when attackers lack access to model gradients. \correct{ Additionally, they examined the transferability of adversarial examples between CNNs and transformers}. Building upon and complementing these efforts, we conduct our evaluations on ImageNet, incorporating both CNN and transformer architectures, and including an ensemble defense strategy that combines robust and standard base models. Our approach further investigates transferability between adversarially trained and standardly trained models. We leverage SoTA robust models to challenge the attacks used to evaluate robustness.

In the field of computational pathology, \cite{ghaffari2022adversarial} examined adversarial robustness with a focus on CNN and transformer models. This work addressed certain adversarial attacks (including Fast Adaptive Boundary (FAB) \cite{croce2020minimally}, and Square \cite{andriushchenko2020square}), as well as adversarial training using (PGD) with Dual-Batch Normalization (DBN). Our work extends this line of inquiry by incorporating adversarially trained SOTA models and a more comprehensive suite of black-box attacks, providing additional context for these attacks by increasing their number and offering a more challenging confrontation against SoTA defenses.

\newtext{Zheng et al.\cite{zheng2025blackboxbench} presented  the BlackboxBench benchmark, which offers a comprehensive study for evaluating black-box adversarial attacks. Specifically, they implemented 59 black-box attack methods, including 29 query-based  and 30 transfer-based algorithms. They conducted evaluations on two datasets 
(CIFAR-10 and ImageNet) and examined four attack settings (targeted and untargeted  attacks under \(L_2\) and \(L_\infty\) norms). They explored both regular and  adversarially  trained models . Our work expands the evaluation of target and surrogate models by assessing black-box attacks with the most recent SoTA adversarially trained models from the RobustBench leaderboard. We revisit the role of alignment between surrogate and target models, confirming that training-strategy compatibility enhances transferability, a trend also observed in BlackboxBench, where we updated this analysis with SoTA defenses both as surrogates and as targets. }

A recent line of research investigated preprocessing and randomness injection defenses against black-box attacks. \cite{qin2021random} proposes adding Gaussian noise to queries to disrupt gradient estimation in black-box attacks, combining it with Gaussian-augmentation fine-tuning for improved robustness-accuracy trade-offs; \cite{chen2022adversarial} proposed the Adversarial Attack on Attackers (AAA) defense. It introduces output logit perturbation to misdirect score-based attacks while preserving clean accuracy and improving calibration. \cite{nguyen2024understanding} showed that randomizing hidden features provides better robustness than input randomization against query-based attacks, with plug-and-play implementation and minimal accuracy loss. \cite{sitawarin2023preprocessors} demonstrated that unaware preprocessing reduces attack efficacy by 7×, and designed preprocessor-aware attacks that easily overcome such defenses.

\subsection{Explaining Transferability and Robustness in Black-Box Setting}
\label{subsec:explanation}

Various research investigated the factors behind the transferability (or lack-off) of adversarial examples. In our study, we covered each theory with at least one representative method. 
\paragraph{Decision Boundary Similarity \cite{liu2016delving}} Models trained on the same task develop similar decision boundaries. Adversarial perturbations are highly aligned with weight vectors across models. Attacks relying on these mechanisms include PGD and HopSkipJump.
\paragraph{Gradient Similarity \cite{demontis2019adversarial}} Two main factors contribute to transferability: (1) similarity between gradient directions of source and target models, and (2) smoothness of the loss landscape. Higher gradient similarity and lower variance in loss landscapes lead to increased transferability. Representative attacks include MI-FGSM and Variance Tuning (VMI/VNI).
\paragraph{Shared Adversarial Subspaces \cite{tramer2017space}} Adversarial examples span contiguous subspaces of large dimensionality. When different models are trained on the same task, a fraction of their adversarial subspaces overlap. DI-FGSM, Ghost Networks, and LGV use these mechanisms.
\paragraph{Non-Robust Features \cite{ilyas2019adversarial}} Models learn both robust features (perceivable by humans) and non-robust features (imperceptible but statistically correlated with labels). Adversarial examples exploit non-robust features. ADMIX and NES leverage this theory.
\paragraph{Model Complexity/Capacity \cite{wu2020towards}} Model-specific factors, including architecture and capacity, influence transferability. Adversarial examples generated from simpler/shallower models tend to transfer better to complex models (e.g. SGM).
\paragraph{Interaction-Based \cite{wang2020unified}} There is a negative correlation between adversarial transferability and interactions inside adversarial perturbations. Less interaction leads to higher transferability. TI-FGSM falls into this category.
\paragraph{Knowledge Transferability \cite{liang2021uncovering}} Models with high knowledge transfer (e.g., via fine-tuning) exhibit stronger adversarial example transfer. The embeddings trained in TREMBA leverage this knowledge transfer.
\paragraph{Flatness/Manifold \cite{fan2024transferability}} There are conflicting results that higher flatness of adversarial examples enables better cross-model transferability. BASES, through its search algorithm, finds optimal weights to explore the manifold.

Other researchers specifically explored the link between white-box robustness and transferability. \cite{springer2021little} investigated the link between the robustness of the source models and the robustness of the target models for transferable attacks. Their results show as expected that adversarial examples generated using non-robust networks do not transfer to the adversarially trained networks. However, what they consider as robust source models is a simple Resnet50 with vanilla adversarial training, which has since been shown to only be slightly robust \cite{croce2021robustbench}. Our study, on the other hand, leverages state-of-the-art robustification mechanisms and explores a large set of architectures for robust surrogates and targets. Although our initial results match, our study highlights further insights (e.g., using robust sources is actually detrimental when targeting non-robust targets).

\cite{zhang2024does} confirmed the impact of model smoothness and gradient similarity by exploring the impact of weakly adversarial training (that is, training with mildly perturbed examples). The approach used to train the model, including data augmentation, synthetic data, regularization are considered in some of the models of our study. Our work focuses on robustification mechanism using extreme augmentations. All in all, our results are complementary; as they explore "mildly robust models" and we focus on "extremely robust models".

% Building on these foundational studies, our work aims to present a broader and more nuanced understanding of robustness in modern machine learning models, particularly in the context of black-box adversarial threats.

%% file: pages/3.RQ1-V3.tex
\section{Success Rate of Black-Box Attacks Against Adversarial Training}
\label{sec:RQ1}

As preliminaries, we investigate the effectiveness of 13 black-box adversarial attacks against Madry adversarial training \cite{madry2017towards} on a single, relatively simple ResNet50 model \cite{he2016deep}. 
% we investigate the effectiveness of adversarial training against SoTA black-box attacks 
The purpose of these experiments is to check to what extent black-box attacks, typically successful against vanilla models, remain as successful when confronted to adversarially trained models. This preliminary experiment provides us with the first insight into the impact of defenses studied in white-box settings on the effectiveness of black-box attacks.

%We hypothesize that advances in black-box attacks, both SoTA and simple attacks, are less effective against robust models hardened with a simple adversarial training technique.

% SoTA ineffective against adversarially trained models, yielding performance levels comparable to those of simple baseline attacks

% We hypothesize that all the advances in black-box attacks do not bring any benefits against simple robustification of models, compared to old yet simple blackbox attacks..

\subsection{Experimental Setup}
We selected a ResNet50 vanilla target and a ResNet50 robust target that underwent adversarial training. We use the same model architecture to eliminate the variations that could arise from differing model architectures. 
We followed the Robustbench evaluation protocol \cite{croce2021robustbench} and reused the original implementation of each attack: All attacks are untargeted and bound to a $L_{\infty}$ maximum perturbation with a distance of $4/255$\newtext{, unless mentioned otherwise}. We evaluated the Attack Success Rate (ASR) on 5000 examples from the ImageNet validation set, and we used the same image identifiers (IDs) as in RobustBench.

\reformulatedtext{We selected nine transfer-based adversarial attacks and four query-based adversarial attacks to provide a comprehensive evaluation of adversarial techniques, balancing foundational approaches and SoTA. For transfer attacks, our selection emphasizes the diversity of techniques by including methods targeting input transformations (TI, DI), gradient optimization (MI, VNI, VMI, ADMIX), universality (UAP), architectural nuances (GHOST), and geometry (LGV). 
For query-based attacks, we included two  families. First, pure query attacks, which rely solely on feedback from the target model such that their performance is not influence by the choice of the surrogate model , such as (SQUARE) and (SIGNFLIP). Second, hybrid query attacks, which improve query efficiency and performance by leveraging surrogate models with optimization strategies to reduce the number of queries. Specifically, (BASES) modifies the weights assigned to each surrogate based on feedback from the target, effectively searching a small-dimensional space, while (TREMBA) learns a low-dimensional embedding and conducts an efficient search within this space. The details and hyperparameters for all attacks are provided in the Appendix~\ref{appendix:hyp-transfer}.}

%We employ surrogate models that are individually larger than the target model or collectively larger as an ensemble, allowing us to leverage the full capacity of the attack. 
For all single-surrogate attacks, including the five FGSM-based baseline attacks, ADMIX, UAP, GHOST, and LGV, the surrogate is a WideResNet50-2 \cite{zagoruyko2016wide} with standard training.
For TREMBA, the surrogates are an ensemble of four models with standard training: VGG16 \cite{simonyan2014very}, ResNet18, SqueezeNet \cite{iandola2016squeezenet}, and GoogLeNet \cite{szegedy2015going} as provided in their repository.
For BASES, the surrogates are an ensemble of 10 models with standard training: VGG16$_{BN}$, ResNet18, SqueezeNet$_{11}$, GoogLeNet, MnasNet$_{10}$ \cite{tan2019mnasnet}, DenseNet161 \cite{huang2017densely}, EfficientNet$_{B0}$ \cite{tan2019efficientnet}, RegNet$_{Y400MF}$ \cite{radosavovic2020designing}, ResNeXt101$_{32x8d}$ \cite{xie2017aggregated}, ConvNeXt$_{Small}$ \cite{liu2022convnet} \reformulatedtext{ . The models used in this study were obtained from the torchvision library \cite{torchvision2016} and the RobustBench benchmark \cite{croce2021robustbench}}. We excluded the SGM attack from this section because, at the time of this study, its implementation supports only ResNet and DenseNet surrogates, while our setup uses WideResNet50-2.

\subsection{Results}

\begin{figure*}[ht]
    \centering
    \includegraphics[width=\linewidth]{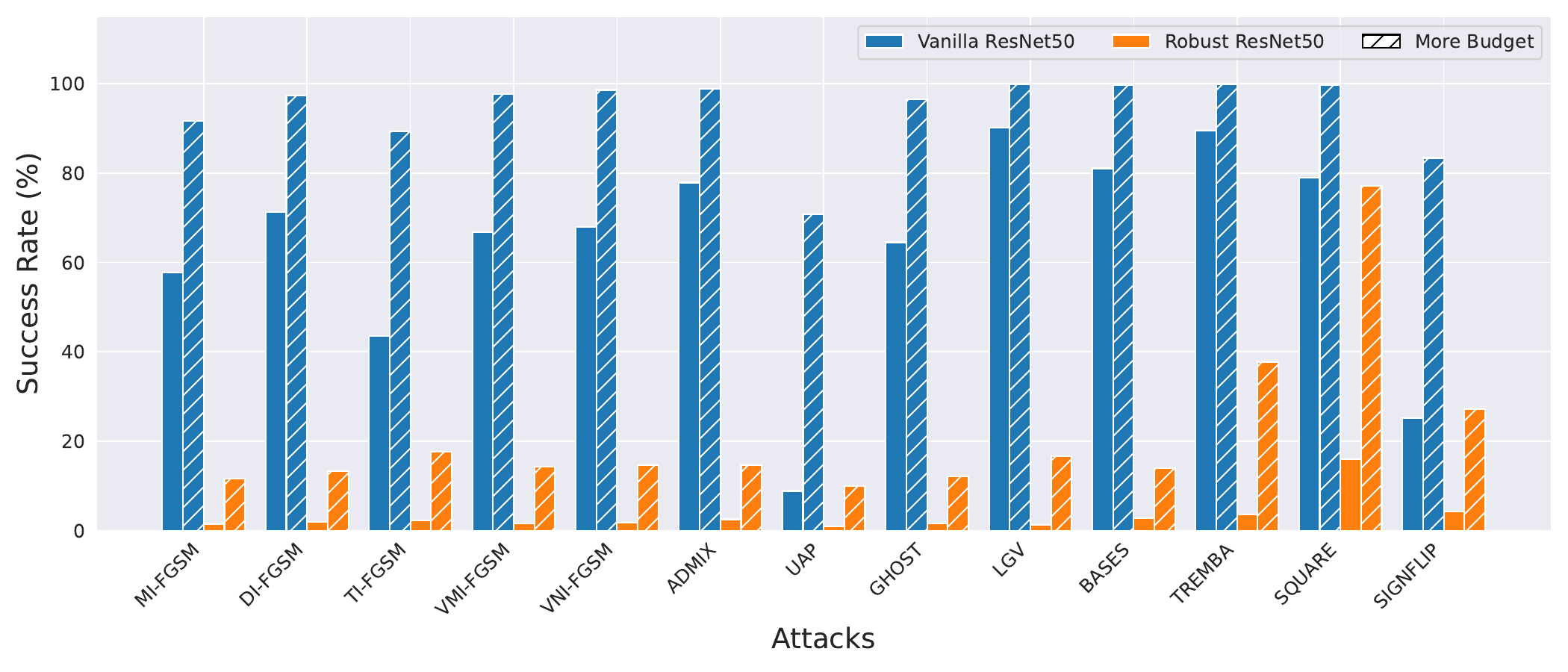}
    \caption{\correct{The blue bars  show the success rates and standard deviations for the vanilla ResNet50 model, while the orange bars show the results for the robust ResNet50 model. Bars in solid color are with attack distance $\epsilon=4/255$, while dashed bars are with attack distance $\epsilon=16/255$}}
    \label{fig:RQ1}
\end{figure*}

\reformulatedtext{We evaluated the ASR of the black-box attacks against standard and adversarially trained Resnet-50 models. We illustrate the ASR using bars  in Figure \ref{fig:RQ1}.
Under a restrictive perturbation budget of 4/255, all attacks except UAP and SIGNFLIP achieve more than 40\% ASR against the undefended model, with the most recent attacks LGV, BASES and TREMBA  reaching 90.11\% ± 0.12, 81.08\% ± 0.49, and 89.56\% ± 0.09, respectively. However, all black-box attacks fail against the robustified model with an ASR of less than 17\%. To rule out query-budget limitations, we quadruple the budgets of the BASES and TREMBA hybrid attacks. Even with this increase, both attacks still perform far worse against the adversarially trained ResNet50 than against the vanilla model (see Appendix \ref{sec:quadruple}).}
\\
\newtext{When the perturbation budget is increased to 16/255 and the query budget for query-only attacks is doubled, black-box attacks can achieve  higher ASR. Under these settings, the performance gap between the undefended and defended models narrows, with the Square Attack reaching an ASR of 79\%.}

% remaining attacks are attacks designed to improve tranferability of gradient based attacks

\subsection{\reformulatedtext{Parameter Sensitivity Analysis}}
\label{sec:rq1-ablation}

\begin{figure*}[ht]
    \centering
    \includegraphics[width=\linewidth]{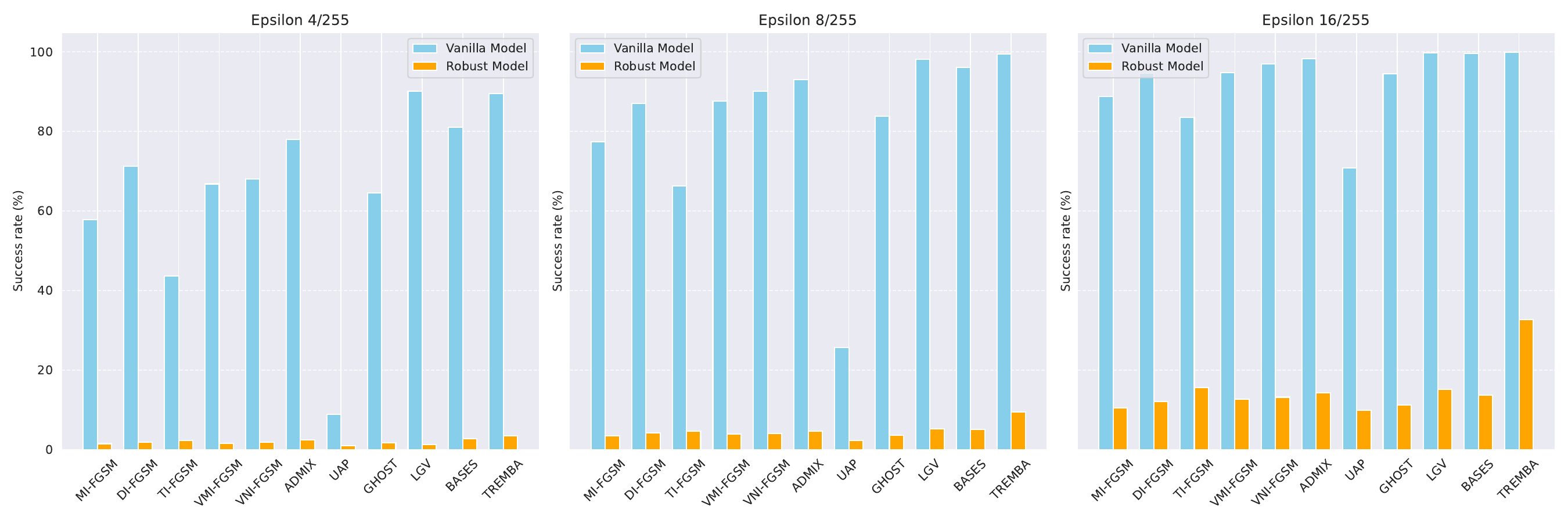}
    \caption{{\reformulatedtext{Sensitivity study of the impact of the perturbation budget $\epsilon$.}}}
    \label{fig:epsilon_ablation}
\end{figure*}

\reformulatedtext{We study in the following the impact of increasing perturbation and computation budgets. In Fig.\ref{fig:epsilon_ablation} we present the results for the increasing $\epsilon$ budgets, and report the exact performance with increased number of iterations and queries in Appendix \ref{sec:eps_budget}. Our results confirm that significantly increasing the perturbation budget does not lead to a collapse of robustness of adversarially trained models and that TREMBA is the only attack with a success rate increasing over 30\% for extremely large budgets. The results in Table \ref{tab:iteration_results_eps} show that increasing iterations has negligible impact, and Table \ref{tab:query_results_eps} that increasing the number of queries has also negligible impact.}

\begin{table}[h!]
\caption{\reformulatedtext{Parameter study of the impact of the iteration budget.}}
\begin{center}
\begin{tabular}{lccc}

\textbf{Attack} & \textbf{Iteration=N} & \textbf{Iteration=2N} & \textbf{Iteration=5N} 
\\
\midrule
MI-FGSM & 88.79 $\rightarrow$ 10.52 & 91.63 $\rightarrow$ 11.61 & 93.79 $\rightarrow$ 12.67 \\
DI-FGSM & 94.55 $\rightarrow$ 12.11 & 97.32 $\rightarrow$ 13.30 & 99.17 $\rightarrow$ 14.38 \\
TI-FGSM & 83.53 $\rightarrow$ 15.61 & 89.31 $\rightarrow$ 17.61 & 93.45 $\rightarrow$ 18.69 \\
VMI-FGSM & 94.79 $\rightarrow$ 12.74 & 97.73 $\rightarrow$ 14.30 & 99.06 $\rightarrow$ 14.82 \\
VNI-FGSM & 96.94 $\rightarrow$ 13.21 & 98.62 $\rightarrow$ 14.67 & 99.45 $\rightarrow$ 14.98 \\
ADMIX & 98.28 $\rightarrow$ 14.24 & 98.90 $\rightarrow$ 14.58 & 99.43 $\rightarrow$ 15.02 \\
UAP & 70.87 $\rightarrow$ 9.93 & 70.78 $\rightarrow$ 9.93 & 72.38 $\rightarrow$ 10.02 \\
GHOST & 94.40 $\rightarrow$ 11.27 & 96.48 $\rightarrow$ 12.21 & 98.41 $\rightarrow$ 13.39 \\
LGV & 99.66 $\rightarrow$ 15.16 & 99.92 $\rightarrow$ 16.61 & 99.97 $\rightarrow$ 16.39 \\
\end{tabular}

\label{tab:iteration_results_eps}
\end{center}
\end{table}

\begin{table}[h!]
\caption{{\reformulatedtext{Parameter study of the impact of number of query budgets.}}}
\label{tab:query_results_eps}

\begin{center}
\begin{tabular}{llcc}

\textbf{Model} & \textbf{Attack} & \textbf{Query=N} & \textbf{Query=2N} 
\\
\midrule
Vanilla & BASES & 99.56 & 99.66 \\
Vanilla & TREMBA & 99.84 & 99.90 \\
Robust & BASES & 13.76 & 14.05 \\
Robust & TREMBA & 32.67 & 37.79 \\
\end{tabular}
\end{center}
\end{table}

\begin{Insight}[boxed title style={colback=pink}]
\reformulatedtext{The effectiveness of black-box attacks drastically decreases against a simple adversarially trained model, with the strongest attack (Square) seeing its ASR drop from 79.04\% to 16.06\%.}
\end{Insight}

% \begin{Insight}[boxed title style={colback=pink}]
% SoTA Black-box attacks are ineffective against adversarially trained models and achieve similar performances as the simplest baseline attacks.
% \end{Insight}

%% file: pages/4.RQ2-V3.tex
\section{Effectiveness of White-Box Defenses Against Black-Box Attacks}
\label{sec:RQ2}

\reformulatedtext{Our first insight demonstrates that simple Madry adversarial training is an effective defense against most SoTA black-box attacks under a restrictive attack budget. We explore in this section whether improved defenses manage to also improve against SoTA black-box attacks in all the settings. These new defenses have been designed to overcome AutoAttack, the SoTA white-box attack, and we are the first to confront them to SoTA black-box attacks.}

%Given that even the most recent advances in black-box attacks remain ineffective against Madry adversarial training, we investigate whether recent innovations in defense mechanisms further reduce the effectiveness of black-box attacks. In particular, we want to observe whether these defenses designed against white-box attacks (viz. AutoAttack) and established benchmarks (viz. Robustbench), generalize to black-box attacks. A positive answer to these questions would maintain (and even raise) the motivation to develop defenses in white-box settings, whereas a negative answer would raise an urgent necessity of investigating specific defenses against black-box attacks.

%Indeed, AutoAttack is a commonly used benchmark for assessing empirical robustness in a white-box setting. However, there is a concern that defenses might be focusing on defeating AutoAttack's unique strategies, resulting in overfitting.

%We hypothesize that leading defenses against AutoAttack will also be more effective against black-box attacks compared to simpler defenses; due to the comprehensive nature of AutoAttack. We hypothesize that improved defenses against AutoAttack do not leave blind spots that could be exploited by SoTA black-box attacks. 

\subsection{Experimental Setup}
\reformulatedtext{We followed the evaluation protocol of the first study on nine robust models from Robustbench. 
We selected models that were (1) peer-reviewed, (2) compatible with all the attacks (for instance Skip connections to run GHOST attacks), and (3) available in a diverse set of architectures (transformers , convolutions), size (Base and Large variants), and defense mechanisms (ensembling, synthetic data augmentation, downsampling layers).
Our selection strategy allows to study the impact of defense mechanisms as a whole, and the impact of size and architecture within the same defense mechanism. We provide more details in Appendix \ref{appendix:defenses}.}
%We ensured to have an even distribution across the robustness spectrum for the nine targets. 

\reformulatedtext{The most robust model, a Swin-L \cite{liu2024comprehensive} achieves 59.64\% robust accuracy, and the least robust model, a ResNet18 \cite{salman2020adversarially} achieves 25.32\% robust accuracy. Table \ref{tab:models} presents the architecture and the number of parameters of the models.} 

\reformulatedtext{These nine models have been defended using different mechanisms. Mechanism A involves adversarial training with extensive data augmentation, regularization, weight averaging, and extensive pretraining. Mechanism B involves non linear ensemble of two models, one defended with adversarial training and one vanilla model with standard training. Mechanism C mitigates adversarial examples using downsample convolutional layers before subsequent network layers. All models with standard adversarial training are denoted with defense D.}

%We kept the same surrogate ensemble for BASES and TREMBA, and we used the surrogate model ResNet50 with standard training for single-surrogate attacks. In addition to the selected black-box attacks, we incorporated the SGM attack due to its compatibility with the ResNet50 surrogate model.

\begin{table*}[ht]
    \centering
     \caption{\reformulatedtext{Robust models included in our evaluation and their configurations. Defense A: Adv training with large data augmentation, regularization, weights averaging, pretraining; Defense B: Non linear ensemble of two base models robust and vanilla; Defense C: Downsample convolutional layers before subsequent network layers; Defense D: Standard adversarial training}}
    %\resizebox{\textwidth}{!}{%
    \begin{tabular}{lcccccc}
     \textbf{Rank} & \textbf{Reference} &  \textbf{Architecture} & \textbf{Robust Accuracy} &  \textbf{Defense} &  \textbf{Architecture Type} & \textbf{Parameters} \\
    \midrule
    1 & \cite{liu2024comprehensive} & Swin-L \cite{liu2021swin} & 59.56 \% & A & Transformer &  196.53M \\ 
    2 & \cite{bai2024mixednuts} & ConvNeXtV2+Swin-L \cite{liu2022convnet,liu2021swin} & 58.50 \% & B & Transformer \& Convolution & 394.49M \\ 
    3 & \cite{liu2024comprehensive} & ConvNeXt-L \cite{liu2022convnet} & 58.48 \% & A & Convolution &  197.77M \\ 
    5 & \cite{liu2024comprehensive} & Swin-B \cite{liu2021swin} & 56.16 \% & A & Transformer &  87.77M \\ 
    7 & \cite{liu2024comprehensive} & ConvNeXt-B \cite{liu2022convnet} & 55.82 \% & A & Convolution &  88.59M \\ 
    12 & \cite{singh2024revisiting} & ViT-S+ConvStem \cite{dosovitskiy2020image, xiao2021early} & 48.08 \% & C & Transformer &  22.78M \\ 
    17 & \cite{salman2020adversarially} & WideResNet50.2 \cite{zagoruyko2016wide} & 38.14 \% & D & Convolution &  68.88M \\ 
    18 & \cite{salman2020adversarially} & Resnet50 \cite{he2016deep} & 34.96 \% & D & Convolution  & 25.56M \\ 
    21 & \cite{salman2020adversarially} & Resnet18 \cite{he2016deep} & 25.32 \% & D & Convolution & 11.69M \\ 
    \end{tabular}%
    %}
   
    \label{tab:models}
\end{table*}

\subsection{Results}

\begin{figure*}
    \centering
    \adjustbox{max width=\textwidth}{%
    
    \includegraphics[width=\linewidth]{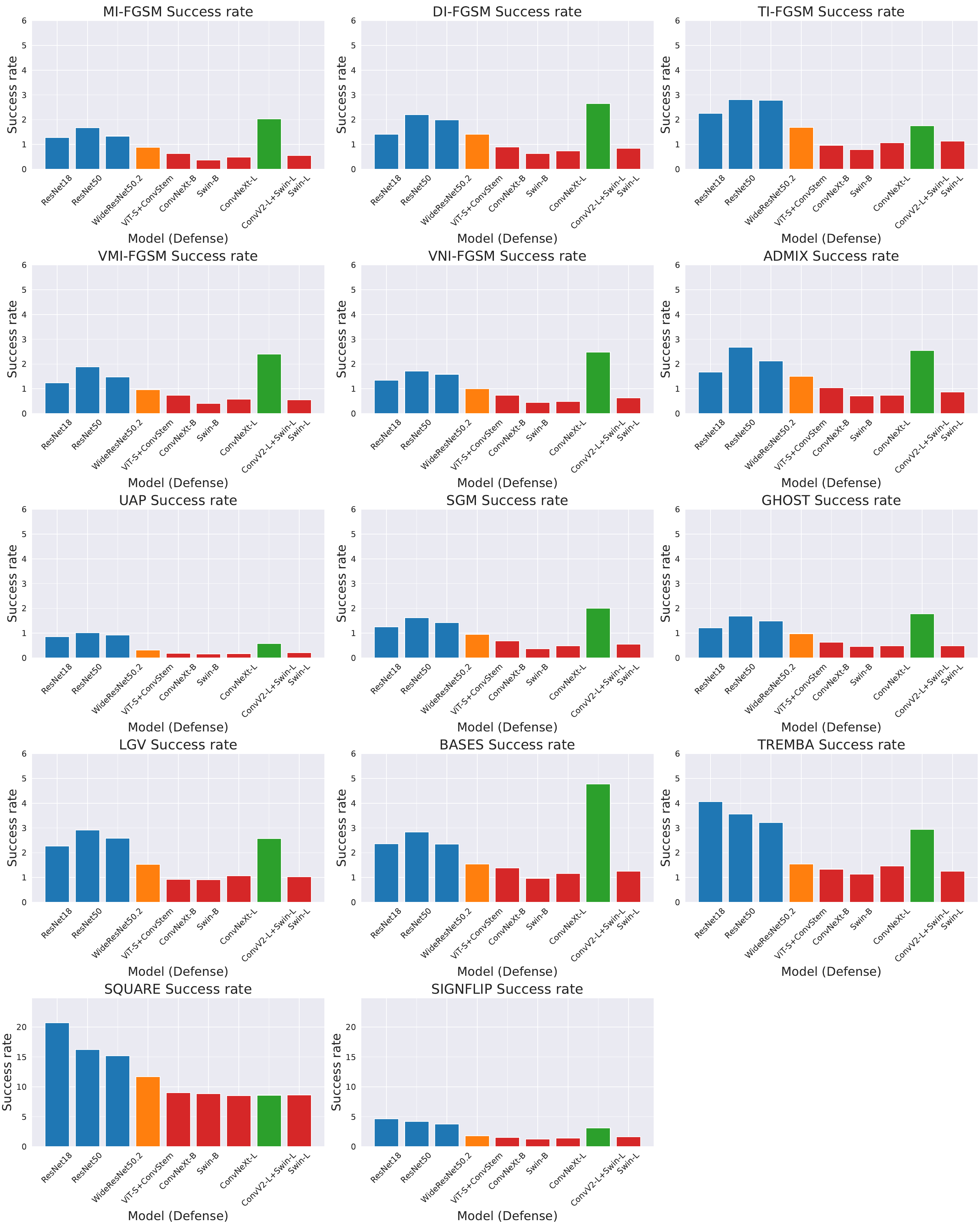}
    }
    
    \caption{\reformulatedtext{Success rate of blackbox attacks against SoTA defenses.}}
    \label{fig:RQ2}
\end{figure*}

We study the performance of defenses against black-box by considering the impact of different defense strategies, the size of the models, and the architecture of the models. We also analyze these results by discussing the relationship between AutoAttack success rates and black-box attack success rates.
% We, next, investigate specifically the impact of the different defense mechanisms on black-box attacks and factors on the success rate of blackbox attacks 

\reformulatedtext{\paragraph{Impact of defense mechanism}
We report in Figure \ref{fig:RQ2} the ASR of the 14 black-box attacks against each of the defended models. The models are sorted from left to right according to the Robustbench leaderboard, with stronger defenses against AutoAttack on the right. Models using the same category of defense mechanisms are grouped with the same color. 
%We observe that stronger defenses against AutoAttack lead to stronger defenses against black-box attacks. 
Defenses from the family A (red) outperform all other defenses. 
Defense C (orange) is better than ensembling (green), and vanilla adversarial training (blue).
Surprisingly, the Conv + Swin ensemble model (orange, ranked 2 against AutoAttack) significanly underperforms models supposedly "weaker" according to Robustbench, and is often weaker than old defenses using only standard training. Conv + Swin is weaker than the vanilla adversarial models against MI-FGSM, DI-FGSM, VM-FGSM, VNI-FGSM, SGM, GHOST, and BASES. We hypothesize that this vulnerability arises because black-box attacks often exploit vanilla surrogates, making the ensemble defense susceptible due to the presence of the standard model.}

%An exception to this trend is observed with the Conv + Swin Defense ensemble, ranked \# 2 on Robustbench, and the mixing mechanisms of types A and B. Although ranked second against AutoAttack, it performs similarly to Defense D when subjected to black-box attacks. This is explained by the composition of defense mechanisms A and B, which combines the adversarially trained model from Defense A with a standard model (vanilla), as explained in Section \ref{subsec:defenses}. This hybrid nature introduces a vulnerability: while this defense ranks among the best defenses against AutoAttack, the inclusion of a standard model component causes its performance against black-box attacks to align with that of weaker AutoAttack defenses. This vulnerability arises because black-box attacks often exploit vanilla surrogates, making the defense susceptible due to the presence of the standard model.

\reformulatedtext{These trends remain consistent when increasing the epsilon values from $4/255$ to $8/255$ and $16/255$, as well as when extending the iteration budget from N to 2N and 5N. Similarly, increasing the query budget from $N$ to $2N$ does not alter the observed pattern. The detailed results are presented in Appendix \ref{sec:eps_budget}.}

% We summarize our results in Fig. \ref{fig:RQ2}. The models are sorted following the Robustbench leaderboard (stronger defenses against AutoAttacks are on the right).
% Except for the ensemble Conv+Swin that combines an adversarially trained model and a standard model, stronger defenses against AutoAttack lead to stronger defenses against black-box attacks.

% Besides, all the attacks are weak against the evaluated defenses, and UAP (respectively TREMBA) is the attack with the lowest (respectively highest) success rate. 
%Given that the surrogate for single model attacks is a standard trained Resnet50, the robust Resnet50 from Salman et al. (second from the left in the plots), achieves higher accuracy than other architecture.  

\begin{table}[t]
\centering
\caption{\reformulatedtext{Success rate on Resnet models trained by Salman et al.\cite{salman2020adversarially}.}}
\begin{tabular}{lccc}
\textbf{Attacks} & \textbf{ResNet18} & \textbf{ResNet50} & \textbf{WideResNet50.2} \\ \midrule
AA & 50.53 & 43.84 & 42.87 \\
MI-FGSM & 1.28±0.00 & 1.68±0.00 & 1.34±0.00 \\
DI-FGSM & 1.41±0.02 & 2.20±0.10 & 1.99±0.04 \\
TI-FGSM & 2.26±0.04 & 2.81±0.07 & 2.78±0.10 \\
VMI-FGSM & 1.24±0.05 & 1.89±0.04 & 1.48±0.05 \\
VNI-FGSM & 1.35±0.02 & 1.72±0.03 & 1.59±0.05 \\
ADMIX & 1.68±0.06 & 2.68±0.13 & 2.12±0.08 \\
UAP & 0.86±0.10 & 1.02±0.01 & 0.93±0.07 \\
SGM & 1.25±0.00 & 1.62±0.00 & 1.43±0.00 \\
GHOST & 1.21±0.18 & 1.69±0.05 & 1.49±0.06 \\
LGV & 2.27±0.11 & 2.92±0.01 & 2.58±0.09 \\
BASES & 2.36±0.03 & 2.83±0.04 & 2.35±0.06 \\
TREMBA & 4.06±0.06 & 3.56±0.06 & 3.22±0.06 \\
SQUARE & 20.58±0.21 &  16.05±0.21 & 15.09±0.1\\
SIGNFLIP & 4.6±0.06 & 4.29±0.1 & 3.79±0.06 \\

\end{tabular}
\label{tab:RQ2_salman}
\end{table}

\reformulatedtext{\paragraph{Impact of model size and architecture}
We focus now on models using the same defense mechanism, and compare the impact of the size of model and archtitectures.}

\reformulatedtext{We compare in Table \ref{tab:RQ2_salman} the success rate of the black-box attacks against models defended with standard adversarial training (family D). These models do not benefit from advanced defense mechanisms (e.g., synthetic data augmentation, ...) and only differ in size. 
While scaling up the models from ResNet18 to WideResNet50 lowers AutoAttack (AA) success from $50.53\%$ to $42.87\%$, this pattern does not hold always for black-box attacks. The black-box attacks that suffer from the increase in target size are TREMBA, SQUARE, and SIGNFLIP with the SQUARE attack dropping from $20.58\%$ to $15.09\%$.}

\reformulatedtext{Moving to Defense A trained models (extensive data enhancement, much larger models, and longer trainings), we report the impact of the increase in model size in Table \ref{tab:RQ2_liu}. While increasing the size of the model (ConvNeXt-B $\rightarrow$ ConvNeXt-L, Swin-B $\rightarrow$  Swin-L) decreases the ASR of AutoAttack, it has no effect on black-box attacks. SIGNFLIP ASR marginally decreases from $1.63\%$ to $1.42\%$ on convolution models with increased size, but marginally improves from $1.28\%$ to $1.65\%$ on Swin transformers with increased size.}

%Although larger models with adversarial training tend to achieve greater robustness to AutoAttack, the increase in size does not lead to lower success rates. Similar insights are revealed in Table \ref{tab:RQ2_liu}, where we compare the robustness of two sizes of ConvNext models and SWIN Transformers. Larger models do not consistently demonstrate stronger robustness to adversarial perturbations in black-box settings, although increased size improves robustness against AutoAttack.

% \paragraph{Impact of model architecture.}
% \label{paragraph:archi}

% \cite{liu2024comprehensive} demonstrate that the best robustness against AutoAttack can be achieved using indiscriminately Convnets or Transformers models. Our results in Table \ref{tab:RQ2_liu} confirm that for a fixed size, both architectures demonstrated similar robustness against black-box attacks.  

\begin{table}[ht]
\centering
\caption{\reformulatedtext{Success rate on very large models trained by Liu et al.\cite{liu2024comprehensive}.}}
\begin{tabular}{lcccc}
\textbf{Attacks} & \textbf{ConvNeXt-B} & \textbf{Swin-B} & \textbf{ConvNeXt-L} & \textbf{Swin-L} \\ \midrule
AA & 28.01 & 27.14 & 25.95 & 25.98 \\
MI-FGSM & 0.63±0.00 & 0.37±0.00 & 0.49±0.00 & 0.56±0.00 \\
DI-FGSM & 0.90±0.02 & 0.63±0.03 & 0.74±0.09 & 0.85±0.04 \\
TI-FGSM & 0.97±0.02 & 0.79±0.07 & 1.07±0.03 & 1.14±0.08 \\
VMI-FGSM & 0.74±0.02 & 0.41±0.01 & 0.58±0.03 & 0.55±0.01 \\
VNI-FGSM & 0.74±0.04 & 0.45±0.02 & 0.49±0.02 & 0.64±0.04 \\
ADMIX & 1.04±0.04 & 0.72±0.02 & 0.74±0.01 & 0.87±0.00 \\
UAP & 0.19±0.04 & 0.16±0.02 & 0.18±0.06 & 0.21±0.04 \\
SGM & 0.69±0.00 & 0.37±0.00 & 0.49±0.00 & 0.56±0.00 \\
GHOST & 0.63±0.03 & 0.46±0.07 & 0.49±0.09 & 0.49±0.04 \\
LGV & 0.92±0.07 & 0.91±0.03 & 1.07±0.03 & 1.03±0.03 \\
BASES & 1.38±0.06 & 0.96±0.02 & 1.16±0.00 & 1.26±0.02 \\
TREMBA & 1.34±0.04 & 1.14±0.03 & 1.47±0.01 & 1.25±0.01 \\
SQUARE & 8.9±0.1 & 8.71±0.12 & 8.56±0.04 & 8.65±0.11 \\
SIGNFLIP &  1.63±0.06 & 1.28±0.03 & 1.42±0.03 & 1.65±0.12 \\

\end{tabular}
\label{tab:RQ2_liu}
\end{table}

\begin{figure}[ht]
    \centering
    \includegraphics[width=\linewidth]{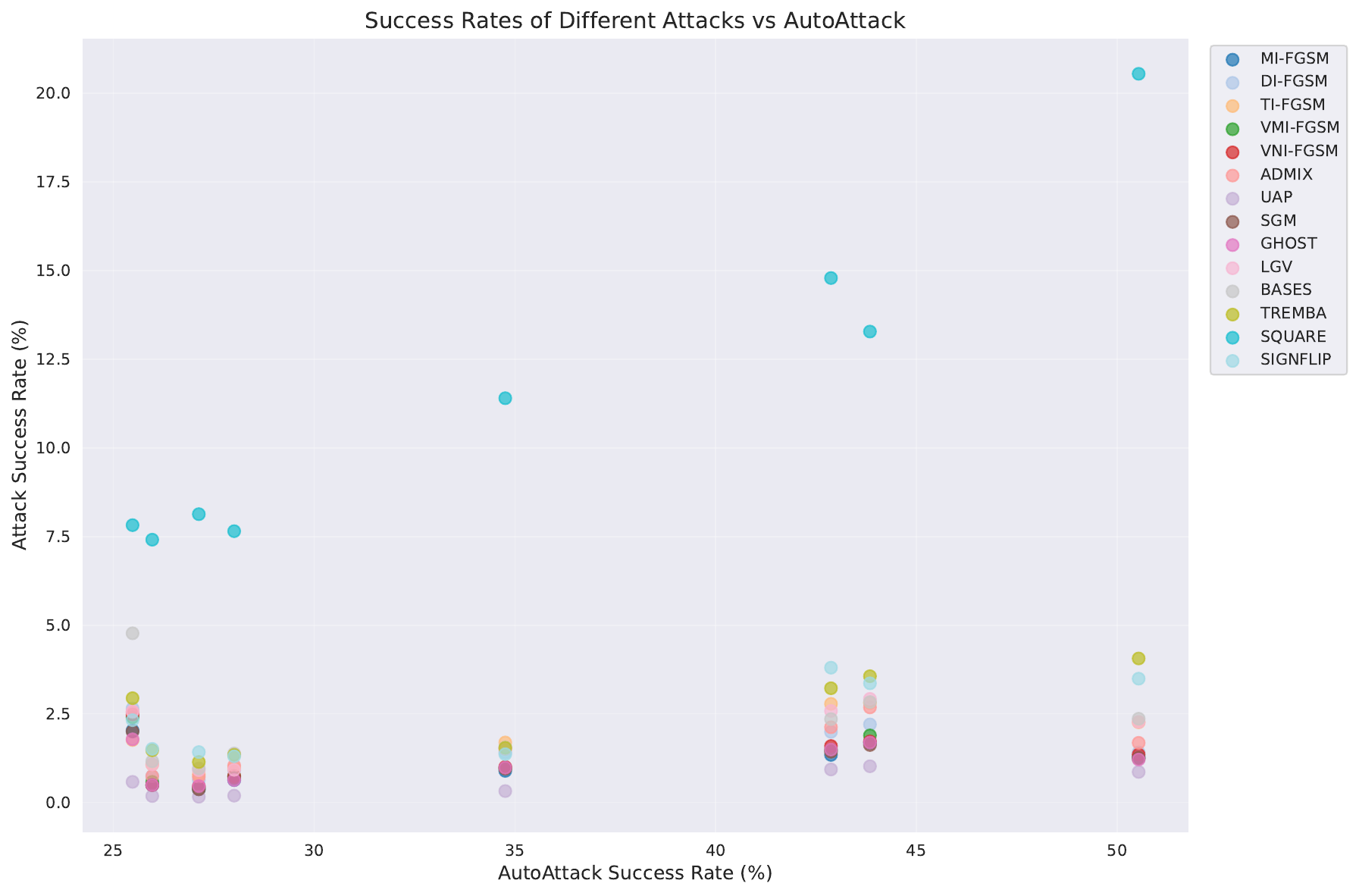}
    \caption{Relation between success rate of AutoAttack and the success rate of black-box attacks.}
    \label{fig:RQ2_autoattack}
\end{figure}

\paragraph{Relationship between AutoAttack success rate and black-box success rate} 
As shown in Figure \ref{fig:RQ2_autoattack}, \reformulatedtext{the scatter plot illustrates the relationship between the ASR of AutoAttack and that of 14 black-box attacks. Each data point represents the ASR on the y-axis of a given black-box attack (indicated by a color) against a specific robust model (determined by its AutoAttack success rate on the x-axis).
% Although the initial ASR of black-box attacks is low (less than 5\%), 
We observe a clear downward trend as robust models improve their effectiveness against AutoAttack, until the best four robust models (with AutoAttack success rate ranging from 25.48\% to 27.14\%). This vulnerability stems from the lack of correlation between model architecture and robustness, as these models all employ defense (A) with diverse sizes and architectures, as shown in Table \ref{tab:models}. }

Furthermore, the best robust model (25.48\% AutoAttack success rate) underperforms compared to the other top robust models against black-box attacks. This occurs because it uses the ensemble defense (B), which, as discussed earlier, combines a robust model with a standard model, making it vulnerable to black-box attacks that often exploit surrogates resembling standard models.

% where an inflection upwards indicates that the best new defenses (against AutoAttack) are more vulnerable to black-box attacks than their predecessors.  

\begin{Insight}[boxed title style={colback=pink}]
SoTA defenses improve robustness to black-box attacks. However, larger models do not guarantee higher robustness and effective ensemble defense against AutoAttack may be less effective against black-box attacks, especially those using surrogate models similar to the ensemble's components. \newtext{Large data augmentation is the best defense against black-box attacks.}
\end{Insight}

% \begin{Insight}[boxed title style={colback=pink}]
% SoTA defenses against AutoAttack generally lead to higher robustness against black-box attacks. However, there is no link between the size or architecture of the defended model and its robustness to black-box attacks. Besides, the last proposed defenses on Robustbench are starting to overfit to AutoAttack and becoming more vulnerable to black-box attacks.
% \end{Insight}

%% file: pages/5.RQ3-V3.tex
\section{Robust surrogates to Improve Black-Box Attacks}
\label{sec:RQ3}

The results from the previous section demonstrated that models robust against AutoAttack remain relevant (to some extent) against black-box attacks. A critical concern then is whether these robust "defenses" could not be used by the attacker to improve its black-box attack. Indeed, most of the black-box require designing a surrogate and can therefore leverage the recent advances in model defense to strengthen their attacks. We hypothesize that these new defenses raise security risks by providing stronger surrogates to attackers.

\subsection{Experimental Setup}

We adhered to the same protocol as in previous sections. We considered various configurations of surrogate and target models. For the target models, we selected highly robust models (Swin-L, ConvNeXtV2+Swin-L, ConvNext-L from \cite{liu2024comprehensive} and \cite{bai2024mixednuts}), a moderately robust model (ResNet18 from \cite{salman2020adversarially}), and a non-robust model (ResNet50). This way, we can measure the effects of using robust surrogates on targets with top defense (A), a simple defense (D) and no defense.

We used the robust surrogate WideResNet-101-2 from \cite{peng2023robust} for single surrogate attacks, because it is the best defense in Robustbench that supports skip connections (which are essential for the GHOST attack). For ensemble surrogate attacks, we also added the robust surrogate Swin-B from \cite{liu2024comprehensive}, as it is the best robust model available, excluding target models. For single, non-robust surrogate attacks, we used a vanilla WideResNet-101-2 in order to have the same architecture as in the robust surrogate, ensuring fair comparison. For BASES and TREMBA (which use an ensemble of surrogates), we kept the same ensemble of vanilla models as in our previous experiments. We excluded the SGM attack that does not support the WideResNet-101-2 surrogates used in our setup.

\subsection{Results}

% \begin{figure}
%     \centering
%     \includegraphics[width=\linewidth]{figures/RQ3_comparison.pdf}
%     \caption{Success rates of SoTA black-box attacks with vanilla and robust surrogates}
%     \label{fig:RQ3}
% \end{figure}

\begin{figure*}[htbp]
    \centering
    \vspace{-1em}
    \adjustbox{max width=\textwidth, max height=0.45\textheight}{%
    \includegraphics{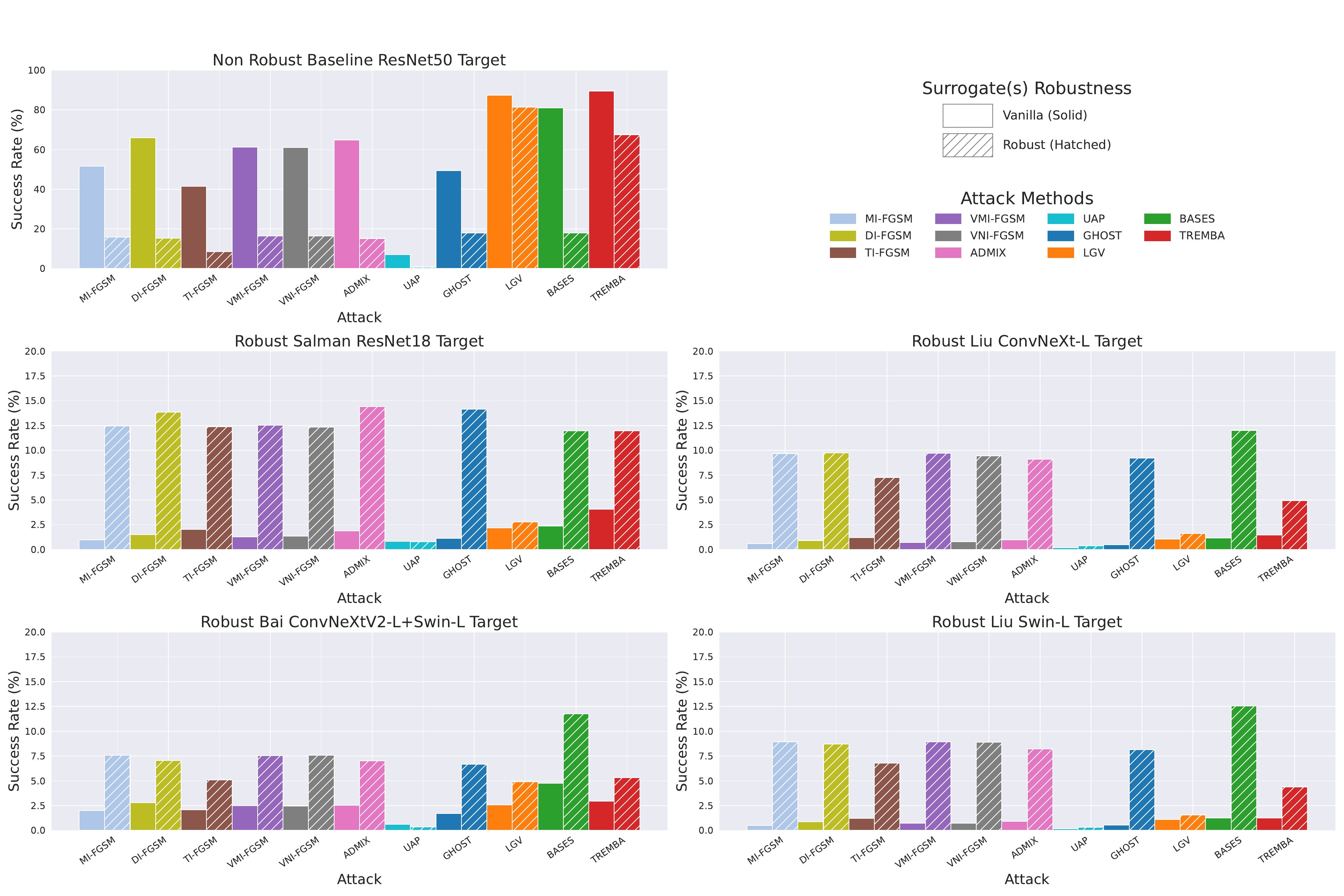}
    }
    \vspace{-1em}
    \caption{Success rates of black-box attacks using vanilla and robust surrogates against a vanilla target and robust targets.}
    \vspace{-1em}
    \label{fig:RQ3-arranged}
\end{figure*}

We present in Figure \ref{fig:RQ3-arranged} the success rate of black-box attacks using non-robust surrogates and robust surrogates against a non-robust target and robust targets. %We evaluate a baseline target, namely a Resnet50 models and a set of 4 robust models, including a SwinL from \cite{liu2024comprehensive}, the best robust models on Robustbench. 

\textbf{Non-robust target:} The results clearly show that the use of robust surrogate models is detrimental to the success rates of the attack. Across all black-box attacks, non-robust surrogates consistently outperform robust surrogates, with an average improvement of 35.19 percentage points in attack success rates on the vanilla ResNet50 model. For instance, GHOST improves from 17.95\% to 49.36\%, BASES from 17.94\% to 81.08\%, and TREMBA from 67.54\% to 89.56\%. Even  UAP that had the lowest success rate improves from 0.73\% to 7.11\%. Notably, attacks that leveraged a training phase before the attacking phase (LGV and TREMBA) exhibited a relatively smaller improvement gap compared to other attacks as robust surrogates were already performing well. Specifically, LGV improved from 81.34\% to 87.39\%, while TREMBA increased from 67.54\% to 89.56\%.

% For instance, success rates using non-robust surrogates on non-robust targets reach 49.36\% with GHOST, 81.08\% with BASES, and 89.56\% with TREMBA. DI-FGSM achieve 65.95\%, VMI-FGSM 61.22\%, ADMIX 64.75\%, and MI-FGSM 51.62\% . Even less effective methods like UAP show a reduction in performance with robust surrogates, dropping from 0.71\% to just 0.48\%. Across all attacks, robust surrogates generally exhibit much lower success rates; for example, GHOST drops from 49.36\% to 17.95\%, BASES from 81.08\% to 17.94\%, and TREMBA from 89.56\% to 67.54\% when switching to robust surrogates. Similar trends are observed for attacks like TI-FGSM, which reduces from 41.5\% to 15.08\%, and DI-FGSM, which drops from 65.95\% to 15.28\%. The average improvement across the eleven attacks for the Baseline ResNet50 target model by using non-robust surrogates instead of robust surrogates is 34.84\% percentage points.

\textbf{Robust target:}
By contrast, robust surrogates yield significant improvement in the success rate of black-box attacks when targeting robust models, resulting in an average improvement of 6.42 percentage points in ASR across attacks and target models. Notably, attacks using robust surrogates on top-tier robust models (e.g., Liu ConvNeXt-L and Liu Swin-L) yield up to 15 times higher success rates than their non-robust counterparts. For instance, when using the GHOST attack to target the highly robust Swin-L model, the robust surrogate achieves a success rate of 8.16\%, compared to just 0.53\% with the non-robust surrogate. Similarly, BASES improves its success rate from 1.26\% to 12.55\% against the Swin-L model. An improvement of more than or equal to 2 percentage points in ASR is consistent across all defense mechanisms and all attacks, except LGV and UAP. We hypothesize that the initial training phase of the LGV models, where a high learning rate is used over additional epochs, weakens the model's robustness during the early sampling phase, which contrasts with GHOST networks that maintain stronger robustness after their creation. We explore this hypothesis in Appendix \ref{sec:LGVvsGHOST}.

The results highlight that the choice of surrogate model plays a critical role in the effectiveness of transfer-based black-box attacks, particularly in restricted environments where the robustness of the target is unknown. Robust models, by design, tend to secure blind spot regions in the input space in which vanilla models often fail to predict adversarial examples correctly. As a result, transfer attacks on robust surrogates aim to fool other regions of the input space that are not yet secured by adversarial training, making them better suited for attacking robust models. On the other hand,
transfer attacks on vanilla surrogates focus more on blind spots of the input space that are also likely to be mispredicted by other vanilla target models.

%\newtext{}

\begin{Insight}[boxed title style={colback=pink}]
The choice of the surrogate model, robust or vanilla, affects the success of black-box attacks. Robust surrogates are more effective against robust targets, while vanilla surrogates excel against vanilla targets. Attackers can therefore improve attack effectiveness by adaptively selecting surrogates based on the perceived robustness of the target model. 
\end{Insight}

% \begin{Insight}[boxed title style={colback=pink}]
% SoTA robust models against AutoAttack can be leveraged to improve black-box attacks against robust models, especially in query settings like BASES and TREMBA. Although using a robust surrogate against an undefended model lowers the success rate of black-box attacks, TREMBA remains effective against both defended and undefended targets when trained with robust surrogates.    
% \end{Insight}

%% file: pages/discussion.tex
\section{Discussion}
\label{sec:discussion}

% observations: Fig1 with large budgets, all attacks are >90% on vanilla, and only Signflip/Square (decision boundary sim) / Tremba (knowledge transfer) over 20% on success rate
% with low distance: only Square  over 5% against adv training
\newtext{\subsection{Impact of defense mechanisms on different attack strategies}}

%\newtext{We discuss in the following how our findings on the impact of each adversarial transferability theory relate to standard adversarial training (SAT) and each of the advanced defenses.}

\newtext{Against advanced defenses, all attack mechanisms are disrupted both with $\epsilon=4/255$ and with $\epsilon=16/255$ (\ref{fig:RQ2_eps16}), meanwhile some attacks mechanism remain effective under  standard adversarial training (SAT):
Compared to vanilla models, SAT is sufficient against all our baseline attacks under low $\epsilon=4/255$ budgets. Under higher $\epsilon=16/255$, only attacks that leverage the \textbf{decision boundary similarity} mechanism (Signflip/Square) and the \textbf{knowledge transfer hypothesis} (TREMBA) achieve over 20\% success rate.} %These low success rate are even more degraded with advanced adversarial trainings techniques (A, B, C from Table \ref{tab:mechanisms}).

\newtext{We explain this observation by the lower impact of SAT on these two mechanisms compared to the other mechanisms. Indeed, SAT alone is not sufficient to disrupt the decision boundaries enough when comparing two models with the same architectures, and the same training data (same "knowledge"). The data augmentations in defenses A and the ensembles in defenses B, however, disrupt TREMBA, Square and Signflip by shifting the knowledge and the boundaries. Attacks like SQUARE, which rely on querying the boundary to estimate its position, struggle because the boundary geometry of defenses A becomes fundamentally different from the surrogate's.}

\newtext{Meanwhile, the \textbf{gradient similarity} attacks like MI-FGSM and VMI-FGSM rely on the assumption that the gradient direction is stable and transferable, which is disrupted by SAT (by smoothing the loss landscape). Using extensive data augmentation and model scaling (defenses A), the boundaries, the knowledge and gradients are "decorrelated" compared to the source model. These defenses also disrupt \textbf{shared space} attacks (DI, GHOST, LGV, UAP) that exploit large, contiguous volumes of empty space in the data manifold to generate adversarial - the large-scale data augmentation fill these spaces, and prevent UAP from finding universal directions of attacks.}

\newtext{\textbf{Interaction-based} attacks (TI-FGSM) and \textbf{non-robust feature} attacks (ADMIX) try to find shortcuts, using respectively low-order (pixel-local) interactions and non-robust features. Our results confirm that these shortcuts are fundamentally disrupted but all adversarial training defenses, making them unreliable for transfer attacks. }    

\newtext{\subsection{Impact of Adaptive Attacks.}}

\newtext{Our third insight confirms that implementing adaptive attacks by including a robust surrogate when crafting the adversarial examples may not be a suitable approach. Although it improves the effectiveness of the attacks (up to 6.42\%), the final success rate achieves at best 12.5\% using adaptive BASES against the strongest defense (Liu Swin-L). Meanwhile, adapting the attacks using a surrogate with a specific defense decreases the effectiveness of the attack against vanilla models. Therefore the adaptation needed in the black-box setting should be carefully crafted for each target.}

\begin{figure}[ht]
    \centering
    \vspace{-1em}
    \includegraphics[width=\linewidth, trim=0 0 0 4cm, clip]{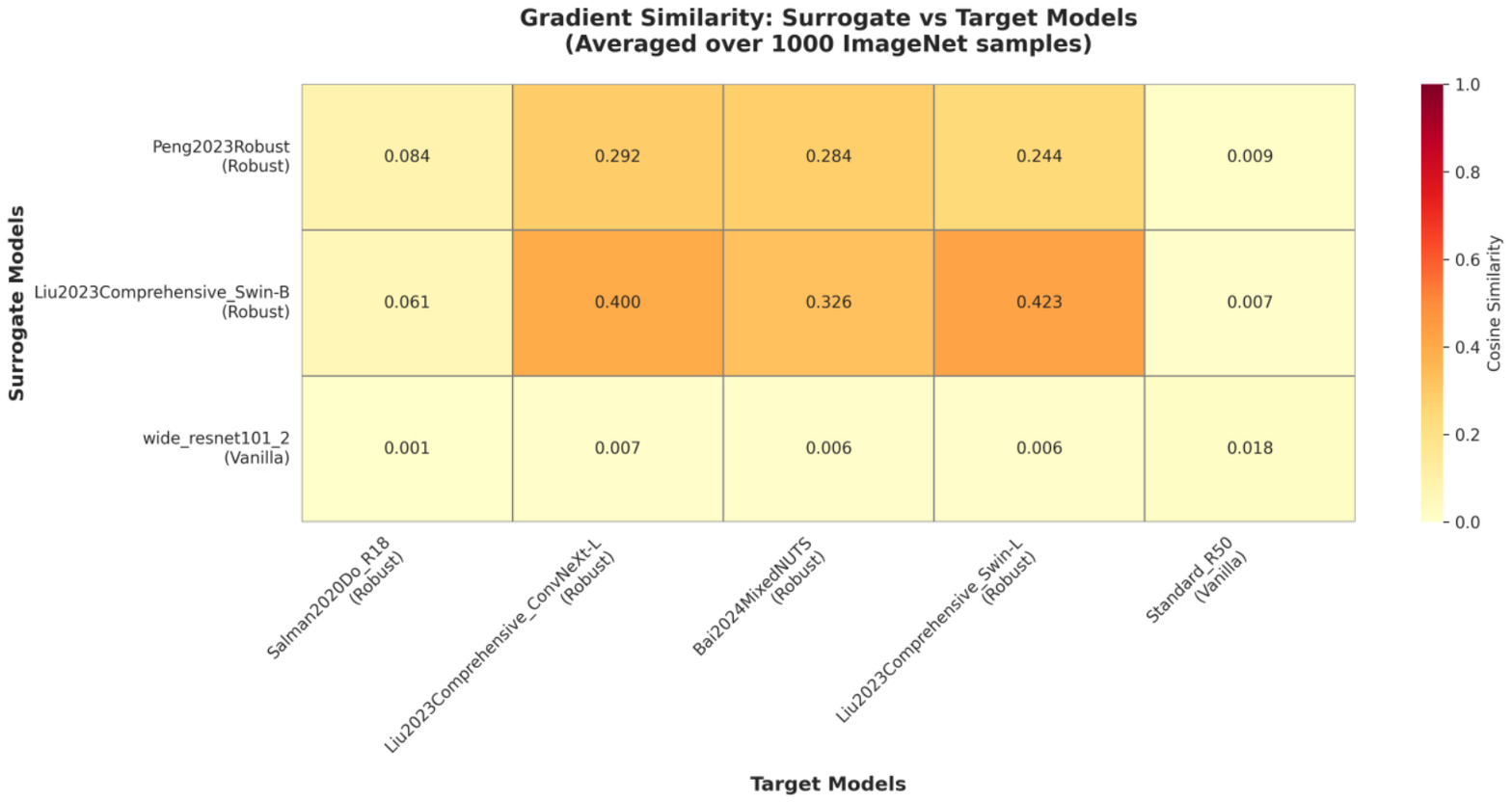}
    \vspace{-4em}
    \caption{Gradient similarity between surrogates and targets over 1000 samples.}
    \vspace{-2em}
    \label{fig:gradient_similarity}
    
\end{figure}

\newtext{To better understand this phenomenon, we measure the gradient similarity across 1,000 ImageNet samples. For each image-label pair, we calculate the gradient of the cross-entropy loss with respect to the input image for both surrogate and target models. We report the cosine similarity between the gradients. This process was repeated for all surrogate-target model pairs, and the similarities were averaged across all samples to produce a similarity matrix in Fig. \ref{fig:gradient_similarity}. Each cell in the matrix represents the average cosine similarity between a surrogate model (rows) and a target model (columns), with values ranging from 0 (orthogonal gradients) to 1 (perfectly aligned gradients). Our results demonstrate that robust targets have a higher gradient alignment with robust surrogates than with vanilla surrogates, even when robust surrogates are from very different architectures and sizes.}

\newtext{\subsection{Transferability Bounds Under Mild Robustness Mechanisms}}

\newtext{Our study focused on SoTA defenses, that have all been designed against strong attacks ($\epsilon=8/255$ in Robustbench). We did not explore the impact of transferability when the models are midly robustified. However, the work promoting mild robustification \cite{trades,fat} was justified only by the tradeoff between robustness and accuracy that was ovserved then. Today, however, very large models and extensive data augmentation allows to robustify the models with agressive attacks and strong mechanisms without degradation of their clean performance (quite the opposite \cite{wang2023better}).}

%and even more with advanced adversarial trainings mechanisms A and B from Table \ref{tab:mechanisms}.       

% Fig 10: Convnext is under performing compared to recent attacks except against TI-FGSM. The strategy A is always the best performing against all attacks

%% file: pages/6-conlusion.tex
% identifying if the level of robsutness is a key factor is important

\section{Limitations and Perspectives}
\label{sec:limitations}
{While our work investigated a large set of black-box attacks and robust models, we mostly focused on extremely robust models. \cite{zhang2024does} for instance focused on mildly perturbed examples and uncovered their interactions with simple adversarial training. Expanding our study of transfer-based attacks to consider varying levels of attack strength and robustness levels between surrogate and target models would also help to deepen our understanding of transferability dynamics. In particular, considering randomness in the pre-processing, and defenses including obfuscation and randomness.}

{In addition, while this study focused on image classification tasks, future research could extend the analysis to other tasks, such as object detection or segmentation, to understand how black-box robustness may vary across applications. These regression tasks raise a new challenge of defining acceptable robustness thresholds, given that the output is not binary. Although there is an established benchmark for adversarial robustness for classification tasks, there is no such benchmark for segmentation tasks.} 

{Another limitation is that while we included large transformers models, we did not explore the robustness of foundations models such as GPT~\cite{GPTradford2018improving}, LLaMA~\cite{touvron2023llama}, CLIP~\cite{radford2021learning}, and SAM~\cite{kirillov2023segment}, or their variants. Research on robustification of these models is still in its infancy for a benchmark.}

\section{Conclusion}

{This research investigated the effectiveness of black-box adversarial attacks against SoTA and standard robust models using ImageNet, structured around three series of experiments, uncovering key insights.. Our first findings demonstrate that even advanced black-box attacks often struggle against a relatively simple adversarial defense. This suggests that adversarial training can be a highly effective baseline defense for real-world systems accessible via APIs. Consequently, understanding the limitations of current black-box attacks against robust models is crucial for developing more effective attack strategies. Based on our second experiment, we found that defense mechanisms optimized against AutoAttack often generalize effectively to resist black-box attacks, even though performance differences are less pronounced in black-box settings. This finding supports the relevance of AutoAttack as a benchmark for black-box robustness and suggests that current efforts focused on white-box defenses provide a meaningful degree of robustness in black-box scenarios as well -- though ensemble defenses may not always translate their AutoAttack robustness to black-box robustness, especially against attacks using similar surrogate models as the ensemble. Finally, we observed that selecting a surrogate with a level of robustness aligned with that of the target can significantly impact the success rate of transfer-based attacks. This finding reveals the possibility for strategic selection of surrogate models when planning black-box attacks. Future attacks should consider designing black-box attacks that specifically target SOTA robust models to better reflect real-world defensive capabilities. Additionally, further exploration of why SoTA defenses optimized for AutoAttack are also relevant for black-box robustness could yield valuable insights into cross-attack resilience mechanisms. In conclusion, our study demonstrates that SoTA defenses are notably resilient against black-box attacks, underscoring the importance of developing more targeted attack strategies to effectively challenge these modern robust models. As the field advances, these insights can guide both the creation of more sophisticated black-box attacks and the design of improved defense mechanisms that support the deployment of robust AI systems across a wide range of applications. } 

%% file: pages/appendix.tex
%\section*{Appendix}

%\subsection{Models}

\subsection{Attacks}
\label{appendix:attacks}
\input{pages/appendix-attacks}

\newtext{\subsection{Defenses}}
\label{appendix:defenses}

\newtext{We have implemented defenses selected following three criteria:
\begin{enumerate}
    \item We only considered defenses from Robustbench from publications that were peer-reviewed at the moment of our study.
    \item We have covered the 3 families of effective defenses in the literature (cf Table IV), while including when possible one strong and one average defense of each family and different variants of the same defense using multiple architectures and sizes. 
    \item We only considered models compatible with all the attacks of the benchmark.  Models must satisfy attack-specific architectural requirements to be used surrogate of all attacks (e.g., WideResNet‑101‑2 supports skip connections, essential for GHOST).
\end{enumerate}
}

\newtext{Salman et al. \cite{salman2020adversarially} evaluated different models sizes and adversarial training epsilon budgets, which led to the models denoted in the following \textbf{Resnet18}, \textbf{Resnet50}, and \textbf{WRN50.2}.}

\newtext{Singh et al. \cite{singh2024revisiting} compared transformers and convnets and demonstrated that ViT + ConvStem yields the best generalization to unseen threat models. We refer to this model as  \textbf{ViT-S+ConvStem}.}

\newtext{Bai et al. \cite{bai2024mixednuts} demonstrated that combining SWIN transformers and large convnets in an ensemble leads to even improved robustness and clean accuracy on ImageNet, we refer to this model as \textbf{ConvNeXtV2-L+Swin-L}.}

\newtext{Finally, Liu et al. \cite{liu2024comprehensive} demonstrated through an extensive benchmark that large SWIN transformers with very large data augmentation during adversarial training, regularization, weight averaging and pretraining outperform all previous approaches. They also demonstrated that large convnets with similar robustification mechanisms can achieve similar robustness. We picked from their models two transformers (one small and one large)  \textbf{Swin-B} and \textbf{Swin-L}, and two convnets \textbf{ConvNeXt-B} and \textbf{ConvNeXt-L}.}

\subsection{Hyperparameters of Transfer Attacks}
\label{appendix:hyp-transfer}

Tables \ref{common_hyperparams} and \ref{specific_hyperparams} provide a comprehensive overview of the hyperparameters used for the ten transfer-based adversarial attacks in our experiments. They categorize the key hyperparameters into two groups: common hyperparameters shared across most attacks and attack-specific hyperparameters that define the unique characteristics of each method. Most of the attack hyperparameters were kept at their default values, as specified in the repositories or libraries that provide the implementations of adversarial attacks \cite{kim2020torchattacks, torchattack}.

\paragraph{Common Hyperparameters}
Table \ref{common_hyperparams} provides the default values and explanations for these common hyperparameters, such as the step size (Alpha), momentum factor (Decay), and number of iterations (Steps), which are essential for iterative attacks as they control the perturbation magnitude and the optimization process.

\begin{table*}
\caption{Common hyperparameters for transfer-based attacks.}
\label{common_hyperparams}
\begin{center}
\resizebox{\textwidth}{!}{
\begin{tabular}{lcll}
\multicolumn{1}{c}{\bf PARAMETER}  & \multicolumn{1}{c}{\bf DEFAULT VALUE} & \multicolumn{1}{c}{\bf EXPLANATION} & \multicolumn{1}{c}{\bf APPLIES TO} \\ \hline \\
Alpha              & 2/255 & Step size or update rate for perturbation.                & All attacks except UAP \\
Decay              & 1.0                  & Momentum factor.                                           & All attacks except UAP \\
Steps              & 10                   & Number of iterations.                                      & All attacks except UAP \\
Epsilon (eps)      & 4/255                & Maximum perturbation for adversarial example.             & All attacks             \\
\end{tabular}
}
\end{center}
\end{table*}

\paragraph{Attack-Specific Hyperparameters}
In addition to the shared hyperparameters, each attack incorporates its own unique hyperparameters tailored to its specific mechanisms. Table \ref{specific_hyperparams} summarizes these attack-specific hyperparameters, listing the default values, explanations, and the corresponding attacks to which they apply.

\begin{table*}
\caption{Attack-specific hyperparameters for transfer-based attacks.}
\label{specific_hyperparams}
\begin{center}
\resizebox{\textwidth}{!}{
\begin{tabular}{lcll}
\multicolumn{1}{c}{\bf PARAMETER} & \multicolumn{1}{c}{\bf DEFAULT VALUE} & \multicolumn{1}{c}{\bf EXPLANATION} & \multicolumn{1}{c}{\bf APPLIES TO} \\ \hline \\
Resize Rate                      & 0.9               & Rate at which the input images are resized.               & DIFGSM, TIFGSM           \\
Diversity Prob                   & 0.5               & Probability to apply input diversity.                     & DIFGSM, TIFGSM           \\
Kernel Name                      & Gaussian          & Type of kernel used.                                       & TIFGSM                   \\
Kernel Length                    & 15                & Length of the kernel.                                      & TIFGSM                   \\
Sigma (nsig)                     & 3                 & Radius of the Gaussian kernel.                            & TIFGSM                   \\
N                                & 5                 & Number of samples in the neighborhood.                    & VMI, VNI                \\
Beta                             & 1.5               & Upper bound for neighborhood.                             & VMI, VNI                \\
Prior Type                       & no\_data          & Type of prior used for attack optimization.               & UAP                     \\
Patience Interval                & 5                 & Number of iterations to wait before checking convergence. & UAP                     \\
Gamma                            & 0.5               & Decay factor for gradients from skip connections.         & SGM                     \\
LGV Epochs                       & 5                 & Number of epochs for training the LGV models.             & LGV                     \\
LGV Models Epoch                 & 8                 & Number of models collected per epoch.                          & LGV                     \\
LGV Learning Rate                & 0.1              & Learning rate for the LGV training phase.                 & LGV                     \\
LGV Batch Size                & 256              & Batch size for the LGV training phase.                 & LGV                     \\
Randomized Modulating Scalar     & 0.22              & Drawn from the uniform distribution 1-scalar, 1+scalar.             & GHOST                   \\
Portion                          & 0.2               & Portion for the mixed image.                              & Admix                   \\
Size                             & 3                 & Number of randomly sampled images.                        & Admix                   \\
\end{tabular}
}
\end{center}
\end{table*}

\paragraph{Hyperparameters of Query Attacks}
\label{appendix:hyp-query}

Tables \ref{BASES_hyperparams} and \ref{TREMBA_hyperparams} provides an overview of the hyperparameters used in  BASES and TREMBA, respectively. Most of these hyperparameters were kept at their default values, as specified in their repositories, since our experiments aimed to evaluate their performance under standard configurations. For BASES, key hyperparameters include the ensemble size (N\_wb), the type of fusion for surrogate models (Fuse), and the number of weight updates (Iterw). Similarly, TREMBA, which operates in two phases (training and attacking), relies on hyperparameters such as the number of training epochs (Epochs), the number of attack iterations (Num\_Iters), the number of samples for Natural Evolution Strategy (NES), and latent space parameters like the Gaussian noise standard deviation (Sigma).

% These parameters encapsulate the balance between optimization strategies, computational constraints, and attack performance.

\begin{table*}
\caption{Hyperparameters for the BASES attack.}
\label{BASES_hyperparams}
\begin{center}
\resizebox{\textwidth}{!}{
\begin{tabular}{lcl}
\multicolumn{1}{c}{\bf PARAMETER} & \multicolumn{1}{c}{\bf DEFAULT VALUE} & \multicolumn{1}{c}{\bf EXPLANATION} \\ \hline \\

N\_wb          & 10                                     & Number of models in the ensemble.        \\
Bound          & linf                                   & Perturbation bound norm type.            \\

Iters          & 10                                     & Number of inner iterations.              \\
Fuse           & loss                                   & Fuse method, e.g., loss or logit.         \\
Loss\_name     & CW                                     & Loss function used for optimization.      \\
Algo & PGD & White-box algorithm used for the perturbation machine \\
X              & 3                                      & Factor to scale the step size alpha.      \\
Learning Rate (lr) & 0.005                              & Learning rate for weight updates.         \\
Iterw          & 20                                     & Number of iterations to update weights.   \\
\end{tabular}
}
\end{center}
\end{table*}

\begin{table*}
\caption{Hyperparameters for the TREMBA attack.}
\label{TREMBA_hyperparams}
\begin{center}
\resizebox{\textwidth}{!}{
\begin{tabular}{lcl}
\multicolumn{1}{c}{\bf PARAMETER} & \multicolumn{1}{c}{\bf DEFAULT VALUE} & \multicolumn{1}{c}{\bf EXPLANATION} \\ \hline \\

Epsilon (eps)       & 8/255 (Training Phase), 4/255 (Attacking Phase)          & Perturbation bound magnitude.            \\
Learning Rate (G) & 0.01 (Training Phase), 5.0 (Attacking Phase)                   & for gradient updates, for latent updates     \\

Momentum       & 0.9 (Training Phase), 0.0 (Attacking Phase)                       & for SGD, for attack on the embedding space.           \\

Num\_images       & 49k (Training Phase)                       & Size of the training set           \\

Epochs         & 500                                                & Number of epochs for Training.            \\
Schedule       & 10                                                 & Epochs between Training learning rate decay.      \\
Gamma          & 0.5                                                & Training Learning rate decay factor.              \\
Margin         & 200.0 (Training Phase), 5.0 (Attacking Phase)                     & Margin for the loss function.            \\

Sample\_Size   & 20                                                 & Number of samples for NES \\
Num\_Iters     & 200                                                & Number of iterations for NES.        \\
Sigma          & 1.0                                                & Standard deviation of the Gaussian noise for NES. \\
Learning Rate Min & 0.1                                             & Minimum learning rate in the Attacking Phase.        \\
Learning Rate Decay & 2.0                                           & Learning rate decay factor in the Attacking Phase.   \\
Plateau\_Length & 20                                                & Number of iterations to monitor the objective function \\
Plateau\_Overhead & 0.3                                             &  Allowed loss threshold to change before the decay \\
\end{tabular}
}
\end{center}
\end{table*}

\begin{table*}
\caption{\newtext{Hyperparameters for the SQUARE attack.}}
\label{SQUARE_hyperparams}
\begin{center}
\resizebox{\textwidth}{!}{
\begin{tabular}{lcl}
\multicolumn{1}{c}{\bf PARAMETER} & \multicolumn{1}{c}{\bf DEFAULT VALUE} & \multicolumn{1}{c}{\bf EXPLANATION} \\ \hline \\

Norm              & Linf            & Norm constraint for the attack ($L_\infty$ or $L_2$).     \\
n\_queries        & 1000            & Maximum number of queries per image.                      \\
n\_restarts       & 1               & Number of random restarts for the attack.                 \\
p\_init           & 0.8             & Initial probability for selecting attack region.          \\
Loss              & margin          & Loss function used for optimization.                      \\
resc\_schedule    & True            & Whether to use rescaling schedule during attack.          \\
\end{tabular}
}
\end{center}
\end{table*}

\begin{table*}
\caption{\newtext{Hyperparameters for the SIGNFLIP attack.}}
\label{SIGNFLIP_hyperparams}
\begin{center}
\resizebox{\textwidth}{!}{
\begin{tabular}{lcl}
\multicolumn{1}{c}{\bf PARAMETER} & \multicolumn{1}{c}{\bf DEFAULT VALUE} & \multicolumn{1}{c}{\bf EXPLANATION} \\ \hline \\

resize\_factor    & 2.0             & Dimensionality reduction rate for low-dimensional space.  \\
max\_queries      & 1000            & Maximum number of queries per image.                      \\
Alpha (initial)   & 0.004           & Initial step size for projection (adaptively adjusted).   \\
Prob (initial)    & 0.999           & Initial probability for sign flipping (adaptively adjusted). \\
Binary search iterations & 10       & Number of iterations for $L_\infty$ binary search initialization. \\
\end{tabular}
}
\end{center}
\end{table*}

\newtext{\section{Computational Budget of Attacks}}

\newtext{Table \ref{model_calls} details the computational budgets of all studied black-box attacks, quantified as the number of model calls required during the attacking phase per entry.}

\begin{table*}
\caption{Computational budgets of black-box attacks based on model calls}
\label{model_calls}
\begin{center}
\resizebox{\textwidth}{!} {
\begin{tabular}{lll}
\textbf{Attack Algorithm} & \textbf{Number of Model Calls} & \textbf{Explanation} \\ \hline \\ 

MIFGSM & Iterations (e.g., 10) & MIFGSM iteratively refines the adversarial image over a number of steps, \\
 &  & each requiring a model call \\\\
DIFGSM & Iterations (e.g., 10) & DIFGSM performs iterative updates with input diversity applied at each step, \\
 &  & requiring a model call per step \\\\
TIFGSM & Iterations (e.g., 10) & TIFGSM performs iterative updates with translation-invariant gradient smoothing \\
 &  & at each step, requiring a model call per step \\\\
VMIFGSM & Iterations * (N + 1) (e.g., 60) & VMIFGSM iterates Steps times. In each iteration, there's one model call for the  adversarial \\
 &  & image and N model calls for the neighboring images to calculate the gradient variance \\\\
VNIFGSM & Iterations * (N + 1) (e.g., 60) & VNIFGSM, similar to VMIFGSM. In each iteration, there is one model call for the  Nesterov and \\
 &  & N model calls for the neighboring images to calculate the gradient variance \\\\
GHOST & Iterations (e.g., 10) & GHOST applies Gaussian noise to the skip connection during each iteration \\
 &  & (one Ghost model per iteration), requiring one model call per step \\\\
LGV & Iterations (e.g., 10) & LGV uses an ensemble of models  (lgv\_nb\_models\_epoch * lgv\_epochs models),\\
 &  & with one model call at each step for each model in the ensemble \\\\
Admix & Iterations * Scale * Size (e.g., 150) & In each step, Admix generates Size admixed images, scales them Scale times, \\
 &  & and then calculates the loss \\\\
UAP & Max Iterations + Evaluation Calls  & The primary loop runs for max\_iter iterations. Additionally, a function is \\
 &  (preparing the perturbation), then 1 & called periodically to evaluate the current perturbation \\\\
TREMBA & (num\_iters + 1) * sample\_size (e.g., 4000) & Tremba iteratively refines the adversarial perturbation within the latent space of a generator network.\\
 &  &  Each iteration involves generating sample\_size perturbations and evaluating their losses \\\\
BASES & iterw * n\_iters * n\_wb (e.g., 2,000) & BASES uses a surrogate ensemble containing n\_wb models and performs an attack \\
 &  & for n\_iters iterations during each query (iterw) \\
 
\end{tabular}
}
\end{center}
\end{table*}

\subsection{Increasing the Computation Budget of Black-Box Attacks}
\label{sec:quadruple}

We are keeping the perturbation budget $\epsilon=4/255$, and quadrupled (multiplied by 4) the budget of the best black-box attacks in section \ref{sec:RQ1} (BASES and TREMBA) against the robust ResNet50 model, to reassure that black-box attacks reduced effectiveness (compared to their effectiveness against the vanilla ResNet50 model) is not due to a limited budget. For BASES, we increased the number of inner iterations of PGD within the perturbation machine (Iters) to 40. For TREMBA, we increased the Number of iterations for NES (Num\_Iters) to 800. 

% To reassure and confirm that black-box attacks struggle against simple adversarial training, and that increasing the black-box attacks budget will not make them as effective as they are against vanilla models, we quadrupled (multiplied by 4 ) the budget for the best black-box attacks in RQ1 which appeared to be BASES and TREMBA

\begin{figure}[ht]
    \centering
    \includegraphics[width=\linewidth]{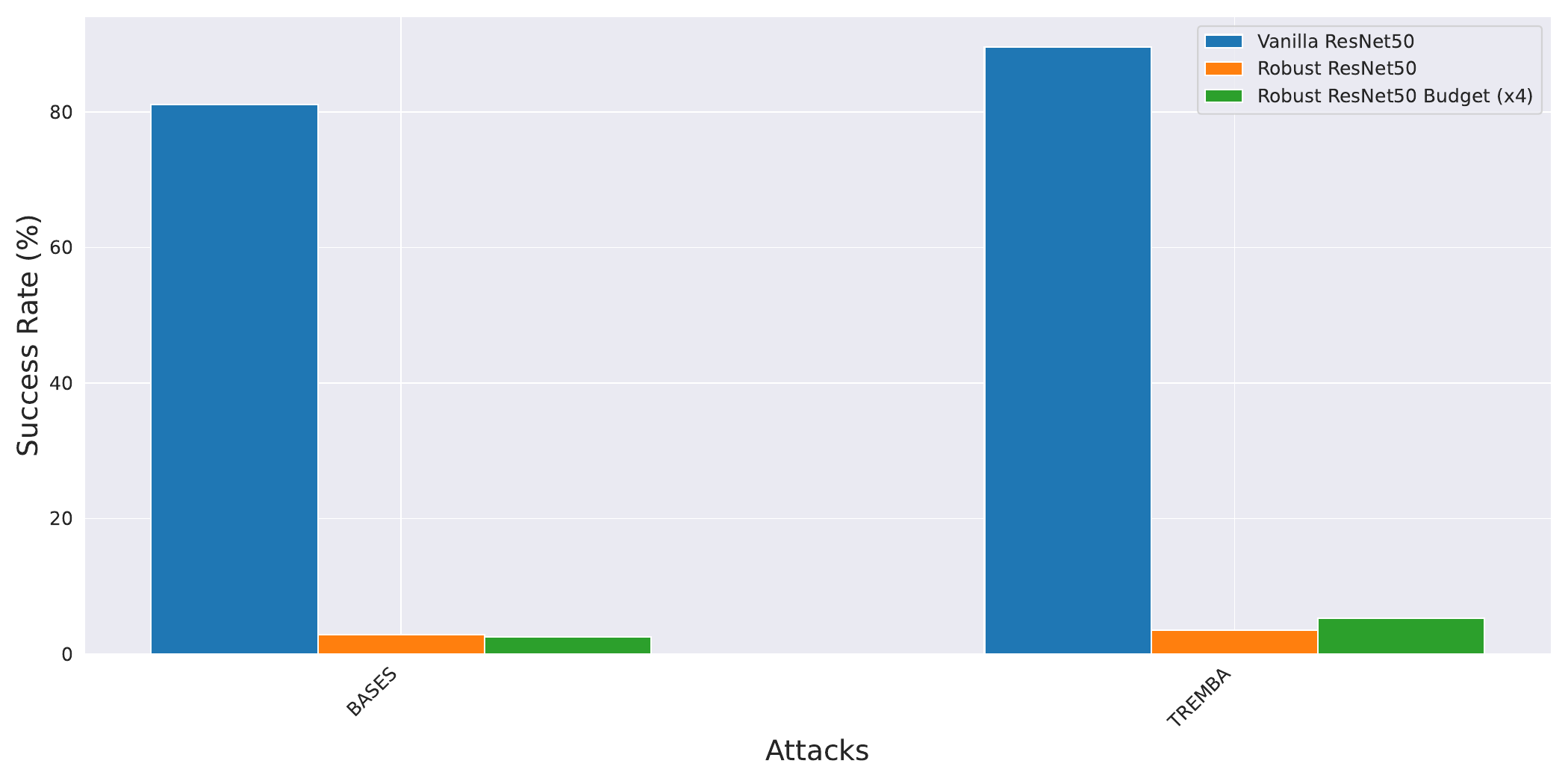}
    \caption{The blue bars show the success rate for the vanilla ResNet50 model, while the orange bars show the results for the robust ResNet50 model, and the green bars show the success rate for the ResNet50 model by quadrupling the budget of the attacks.}
    \label{fig:RQ1_quadruple}
\end{figure}

Figure \ref{fig:RQ1_quadruple} presents the success rates of BASES and TREMBA attacks against the vanilla ResNet-50 model with standard attack budget, and against the adversarially trained ResNet-50 model under both  standard and quadrupled attack budgets. Both attacks underperform against the adversarially trained ResNet50 under quadrupled budgets compared to the vanilla model with standard budgets. For instance, BASES achieves 2.56\% ASR on the robust ResNet50 Budget (x4) versus 81.08\% on the vanilla ResNet50. Similarly, TREMBA's ASR is at 5.24\% on the robust ResNet50 Budget (x4) compared to 89.56\% on the vanilla ResNet50. These findings confirm that our earlier results in section \ref{sec:RQ1} were not a consequence of budget limitations.

\newtext{\section{Increasing the Perturbation Budget of Black-Box Attacks}
\label{sec:eps_budget}}

\newtext{We have increased the perturbation budget to cover $\epsilon \in \{4/255,8/255,16/255\}$. In Table \ref{tab:attack_results_eps}, Table \ref{fast_eps}, and Table \ref{slow_eps}, we only change the perturbation budget $\epsilon$.}

\newtext{
Once we have increased the perturbation budget to $\epsilon=16/255$, we first study the impact of increasing the iteration budget in Table \ref{tab:iteration_results_eps}, and Table \ref{fast_steps}, then, we study the impact of increasing the query budget in Table \ref{tab:query_results_eps}, and Table \ref{slow_steps}.}

\begin{figure*}[b]
    \centering
    \adjustbox{max width=\textwidth, max height=0.45\textheight}{%
    
    \includegraphics[width=\linewidth]{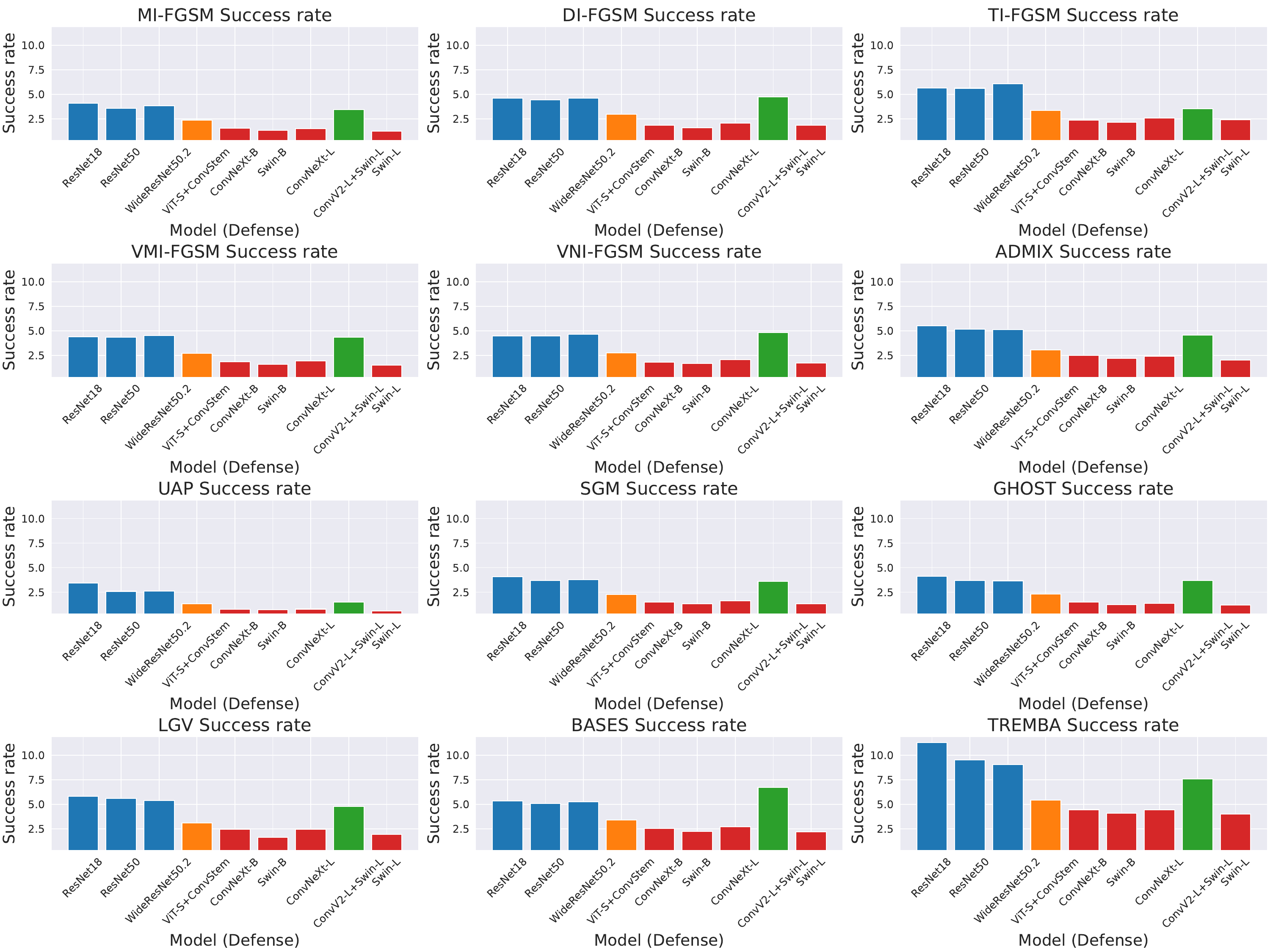}
    }
    
    \caption{Success rate of blackbox attacks against SoTA defenses for epsillon 8/255.}
    \label{fig:RQ2_eps8}
\end{figure*}

\begin{figure*}[b]
    \centering
    \adjustbox{max width=\textwidth, max height=0.45\textheight}{%
    
    \includegraphics[width=\linewidth]{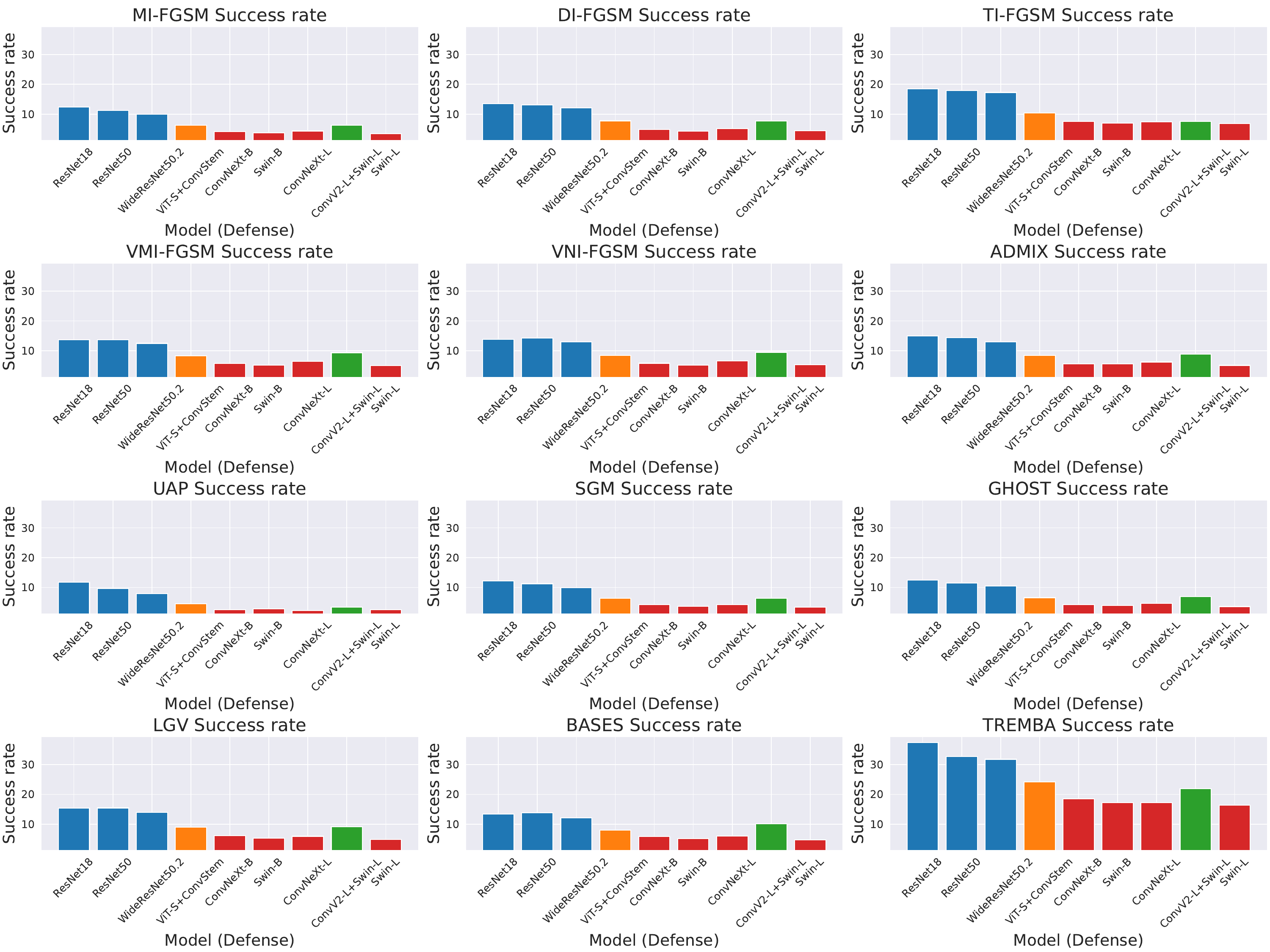}
    }
    
    \caption{Success rate of blackbox attacks against SoTA defenses for epsilon 16/255.}
    \label{fig:RQ2_eps16}
\end{figure*}

\begin{figure*}[b]
    \centering
    \adjustbox{max width=\textwidth, max height=0.45\textheight}{%
    
    \includegraphics[width=\linewidth]{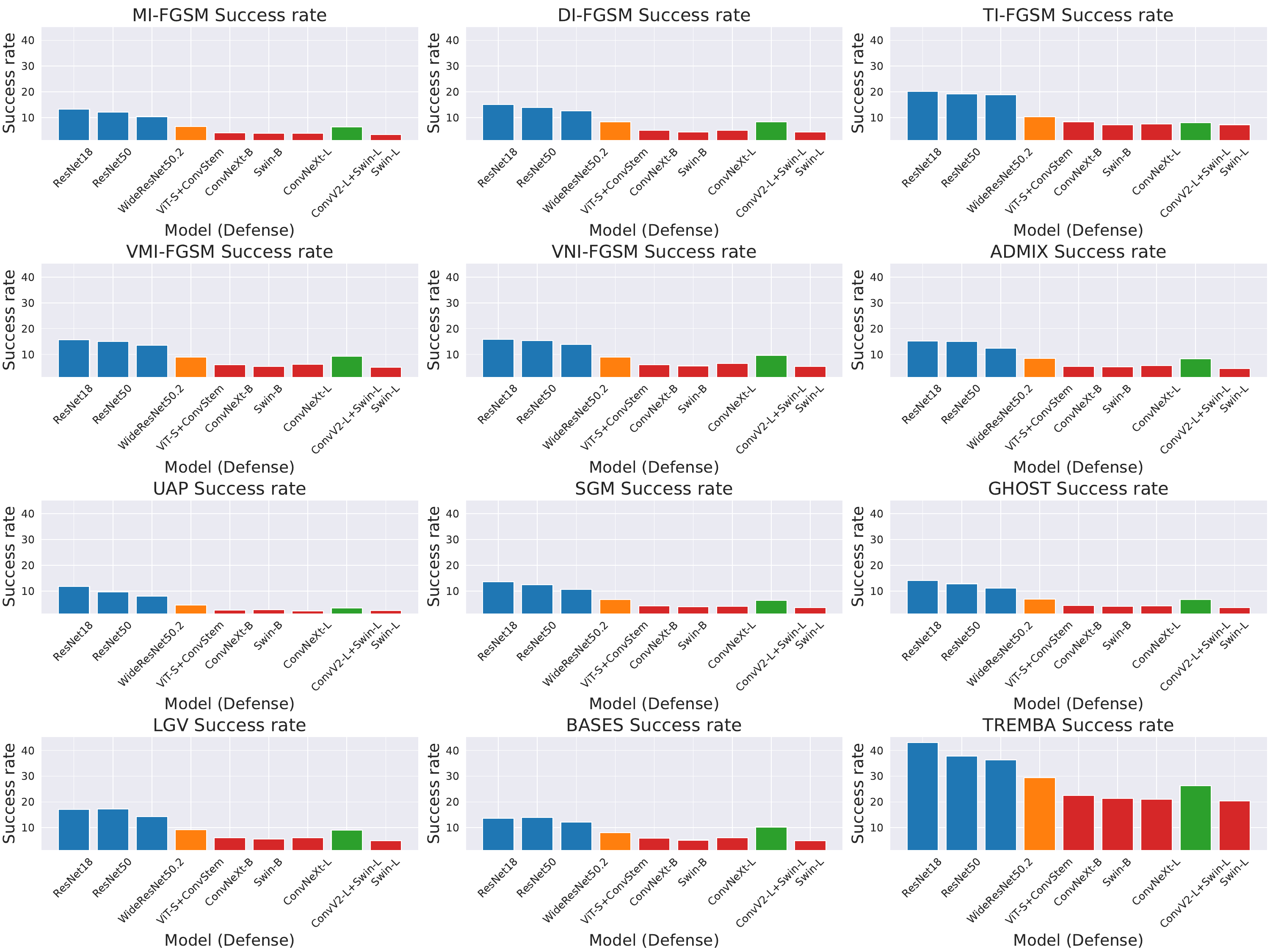}
    }
    
    \caption{\newtext{Success rate of blackbox attacks against SoTA defenses for epsilon 16/255 and iteration=2N.}}
    \label{fig:RQ2_eps16_2N}
\end{figure*}

\begin{figure*}[b]
    \centering
    \adjustbox{max width=\textwidth, max height=0.45\textheight}{%
    
    \includegraphics[width=\linewidth]{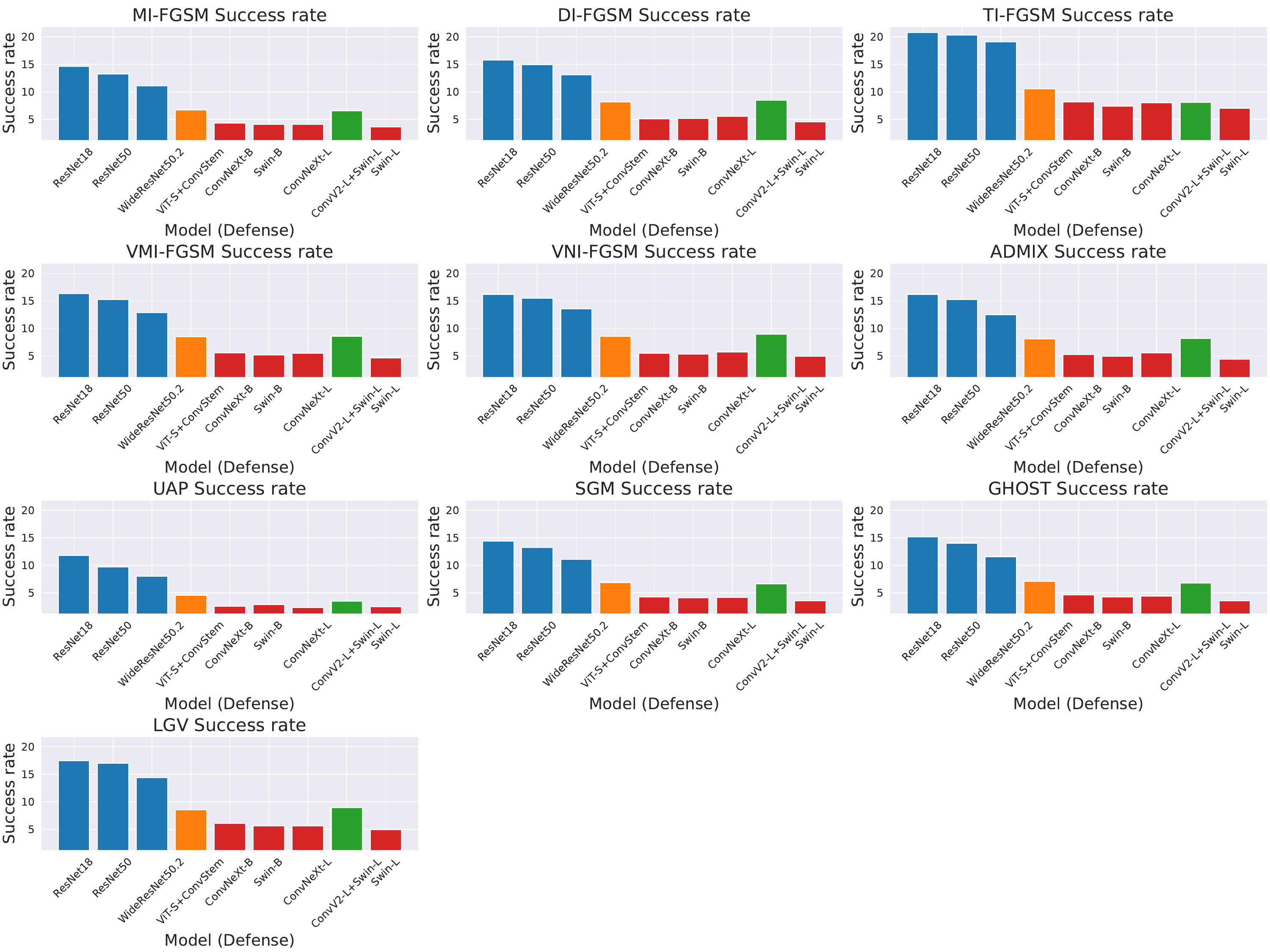}
    }
    
    \caption{Success rate of blackbox attacks against SoTA defenses for epsilon 16/255 and iteration=5N.}
    \label{fig:RQ2_eps16_5N}
\end{figure*}

\newtext{
To further examine the impact of different defense strategies with increasing epsilon values and  budgets, we present the results in the form of figures. Specifically, Figure \ref{fig:RQ2_eps8} and Figure \ref{fig:RQ2_eps16} correspond to epsilon values of 8/255 and 16/255, respectively. Additionally, Figure \ref{fig:RQ2_eps16_2N} for iteration and query budget of 2N and Figure \ref{fig:RQ2_eps16_5N} illustrate results for iteration budgets of 5N.}

\begin{table*}[h!]
\caption{Attack Performance at Different Epsilon Values.}
\begin{center}
\begin{tabular}{lccc}
\textbf{Attack} & \textbf{Epsilon 4/255} & \textbf{Epsilon 8/255} & \textbf{Epsilon 16/255} \\
\hline
MI-FGSM & 57.77 $\rightarrow$ 1.53 & 77.33 $\rightarrow$ 3.53 & 88.79 $\rightarrow$ 10.52 \\
DI-FGSM & 71.31 $\rightarrow$ 1.94 & 87.07 $\rightarrow$ 4.18 & 94.55 $\rightarrow$ 12.11 \\
TI-FGSM & 43.59 $\rightarrow$ 2.30 & 66.35 $\rightarrow$ 4.68 & 83.53 $\rightarrow$ 15.61 \\
VMI-FGSM & 66.79 $\rightarrow$ 1.66 & 87.59 $\rightarrow$ 3.97 & 94.79 $\rightarrow$ 12.74 \\
VNI-FGSM & 68.00 $\rightarrow$ 1.85 & 90.10 $\rightarrow$ 4.09 & 96.94 $\rightarrow$ 13.21 \\
ADMIX & 77.90 $\rightarrow$ 2.54 & 93.04 $\rightarrow$ 4.68 & 98.28 $\rightarrow$ 14.24 \\
UAP & 8.89 $\rightarrow$ 0.98 & 25.70 $\rightarrow$ 2.37 & 70.78 $\rightarrow$ 9.93 \\
GHOST & 64.53 $\rightarrow$ 1.68 & 83.87 $\rightarrow$ 3.59 & 94.40 $\rightarrow$ 11.27 \\
LGV & 90.11 $\rightarrow$ 1.33 & 98.07 $\rightarrow$ 5.25 & 99.66 $\rightarrow$ 15.16 \\
BASES & 81.08 $\rightarrow$ 2.83 & 96.06 $\rightarrow$ 5.09 & 99.56 $\rightarrow$ 13.76 \\
TREMBA & 89.56 $\rightarrow$ 3.56 & 99.45 $\rightarrow$ 9.52 & 99.84 $\rightarrow$ 32.67 \\
\end{tabular}
\label{tab:attack_results_eps}
\end{center}
\end{table*}

\begin{table*}
\caption{Transfer Attack Performance Across Increased Epsilon 4/255, 8/255 and 16/255.}
\label{fast_eps}
\begin{center}
\resizebox{\textwidth}{!} {
\begin{tabular}{lllllllllll}
\textbf{Model} & \textbf{MI-FGSM} & \textbf{DI-FGSM} & \textbf{TI-FGSM} & \textbf{VMI-FGSM} & \textbf{VNI-FGSM} & \textbf{ADMIX} & \textbf{UAP} & \textbf{SGM} & \textbf{GHOST} & \textbf{LGV} \\ \hline \\ 
ResNet18 & 1.28 $\rightarrow$ 4.08 $\rightarrow$ 12.4 & 1.41 $\rightarrow$ 4.61 $\rightarrow$ 13.5 & 2.26 $\rightarrow$ 5.63 $\rightarrow$ 18.49 & 1.24 $\rightarrow$ 4.39 $\rightarrow$ 13.72 & 1.35 $\rightarrow$ 4.5 $\rightarrow$ 13.84 & 1.68 $\rightarrow$ 5.52 $\rightarrow$ 15.01 & 0.86 $\rightarrow$ 3.44 $\rightarrow$ 11.72 & 1.25 $\rightarrow$ 4.08 $\rightarrow$ 12.25 & 1.21 $\rightarrow$ 4.12 $\rightarrow$ 12.48 & 2.27 $\rightarrow$ 5.82 $\rightarrow$ 15.39 \\ 
    ResNet50 & 1.68 $\rightarrow$ 3.56 $\rightarrow$ 11.24 & 2.2 $\rightarrow$ 4.43 $\rightarrow$ 13.11 & 2.81 $\rightarrow$ 5.62 $\rightarrow$ 17.95 & 1.89 $\rightarrow$ 4.37 $\rightarrow$ 13.71 & 1.72 $\rightarrow$ 4.5 $\rightarrow$ 14.21 & 2.68 $\rightarrow$ 5.18 $\rightarrow$ 14.46 & 1.02 $\rightarrow$ 2.59 $\rightarrow$ 9.65 & 1.62 $\rightarrow$ 3.68 $\rightarrow$ 11.27 & 1.69 $\rightarrow$ 3.68 $\rightarrow$ 11.46 & 2.92 $\rightarrow$ 5.59 $\rightarrow$ 15.45 \\ 
    WideResNet50.2 & 1.34 $\rightarrow$ 3.82 $\rightarrow$ 10.02 & 1.99 $\rightarrow$ 4.6 $\rightarrow$ 12.09 & 2.78 $\rightarrow$ 6.06 $\rightarrow$ 17.27 & 1.48 $\rightarrow$ 4.52 $\rightarrow$ 12.5 & 1.59 $\rightarrow$ 4.66 $\rightarrow$ 13.02 & 2.12 $\rightarrow$ 5.13 $\rightarrow$ 12.94 & 0.93 $\rightarrow$ 2.62 $\rightarrow$ 7.95 & 1.43 $\rightarrow$ 3.79 $\rightarrow$ 9.96 & 1.49 $\rightarrow$ 3.67 $\rightarrow$ 10.52 & 2.58 $\rightarrow$ 5.39 $\rightarrow$ 13.92 \\ 
    ViT-S+ConvStem & 0.89 $\rightarrow$ 2.36 $\rightarrow$ 6.26 & 1.41 $\rightarrow$ 2.98 $\rightarrow$ 7.76 & 1.69 $\rightarrow$ 3.34 $\rightarrow$ 10.38 & 0.97 $\rightarrow$ 2.73 $\rightarrow$ 8.37 & 1.0 $\rightarrow$ 2.75 $\rightarrow$ 8.51 & 1.51 $\rightarrow$ 3.09 $\rightarrow$ 8.43 & 0.32 $\rightarrow$ 1.31 $\rightarrow$ 4.56 & 0.95 $\rightarrow$ 2.28 $\rightarrow$ 6.37 & 0.98 $\rightarrow$ 2.31 $\rightarrow$ 6.59 & 1.53 $\rightarrow$ 3.11 $\rightarrow$ 8.93 \\ 
    ConvNeXt-B & 0.63 $\rightarrow$ 1.56 $\rightarrow$ 4.2 & 0.9 $\rightarrow$ 1.85 $\rightarrow$ 4.88 & 0.97 $\rightarrow$ 2.35 $\rightarrow$ 7.5 & 0.74 $\rightarrow$ 1.85 $\rightarrow$ 5.81 & 0.74 $\rightarrow$ 1.82 $\rightarrow$ 5.83 & 1.04 $\rightarrow$ 2.51 $\rightarrow$ 5.6 & 0.19 $\rightarrow$ 0.74 $\rightarrow$ 2.53 & 0.69 $\rightarrow$ 1.5 $\rightarrow$ 4.3 & 0.63 $\rightarrow$ 1.48 $\rightarrow$ 4.22 & 0.92 $\rightarrow$ 2.48 $\rightarrow$ 6.1 \\ 
    Swin-B & 0.37 $\rightarrow$ 1.35 $\rightarrow$ 3.72 & 0.63 $\rightarrow$ 1.59 $\rightarrow$ 4.3 & 0.79 $\rightarrow$ 2.15 $\rightarrow$ 6.93 & 0.41 $\rightarrow$ 1.62 $\rightarrow$ 5.18 & 0.45 $\rightarrow$ 1.7 $\rightarrow$ 5.2 & 0.72 $\rightarrow$ 2.2 $\rightarrow$ 5.55 & 0.16 $\rightarrow$ 0.72 $\rightarrow$ 2.81 & 0.37 $\rightarrow$ 1.33 $\rightarrow$ 3.74 & 0.46 $\rightarrow$ 1.25 $\rightarrow$ 3.95 & 0.91 $\rightarrow$ 1.65 $\rightarrow$ 5.31 \\ 
    ConvNeXt-L & 0.49 $\rightarrow$ 1.5 $\rightarrow$ 4.35 & 0.74 $\rightarrow$ 2.07 $\rightarrow$ 5.18 & 1.07 $\rightarrow$ 2.57 $\rightarrow$ 7.39 & 0.58 $\rightarrow$ 1.97 $\rightarrow$ 6.45 & 0.49 $\rightarrow$ 2.07 $\rightarrow$ 6.56 & 0.74 $\rightarrow$ 2.41 $\rightarrow$ 6.25 & 0.18 $\rightarrow$ 0.78 $\rightarrow$ 2.31 & 0.49 $\rightarrow$ 1.61 $\rightarrow$ 4.28 & 0.49 $\rightarrow$ 1.37 $\rightarrow$ 4.64 & 1.07 $\rightarrow$ 2.44 $\rightarrow$ 5.93 \\ 
    ConvV2-L+Swin-L & 2.03 $\rightarrow$ 3.46 $\rightarrow$ 6.23 & 2.65 $\rightarrow$ 4.75 $\rightarrow$ 7.67 & 1.76 $\rightarrow$ 3.51 $\rightarrow$ 7.52 & 2.4 $\rightarrow$ 4.38 $\rightarrow$ 9.37 & 2.48 $\rightarrow$ 4.85 $\rightarrow$ 9.52 & 2.55 $\rightarrow$ 4.58 $\rightarrow$ 8.83 & 0.58 $\rightarrow$ 1.51 $\rightarrow$ 3.44 & 2.0 $\rightarrow$ 3.61 $\rightarrow$ 6.38 & 1.78 $\rightarrow$ 3.69 $\rightarrow$ 6.9 & 2.57 $\rightarrow$ 4.8 $\rightarrow$ 9.15 \\ 
    Swin-L & 0.56 $\rightarrow$ 1.25 $\rightarrow$ 3.43 & 0.85 $\rightarrow$ 1.84 $\rightarrow$ 4.46 & 1.14 $\rightarrow$ 2.43 $\rightarrow$ 6.84 & 0.55 $\rightarrow$ 1.51 $\rightarrow$ 5.07 & 0.64 $\rightarrow$ 1.74 $\rightarrow$ 5.28 & 0.87 $\rightarrow$ 2.02 $\rightarrow$ 5.12 & 0.21 $\rightarrow$ 0.59 $\rightarrow$ 2.48 & 0.56 $\rightarrow$ 1.31 $\rightarrow$ 3.41 & 0.49 $\rightarrow$ 1.18 $\rightarrow$ 3.61 & 1.03 $\rightarrow$ 1.92 $\rightarrow$ 4.84 \\
\end{tabular}
}
\end{center}
\end{table*}

\begin{table*}
\caption{Transfer Attack Performance Across Different Iteration Counts N, 2N, 5N For Epsilon 16/255.}
\label{fast_steps}
\begin{center}
\resizebox{\textwidth}{!} {
\begin{tabular}{lllllllllll}
\textbf{Model} & \textbf{MI-FGSM} & \textbf{DI-FGSM} & \textbf{TI-FGSM} & \textbf{VMI-FGSM} & \textbf{VNI-FGSM} & \textbf{ADMIX} & \textbf{UAP} & \textbf{SGM} & \textbf{GHOST} & \textbf{LGV} \\ \hline \\ 
    ResNet18 & 12.4 $\rightarrow$ 13.38 $\rightarrow$ 14.63 & 13.5 $\rightarrow$ 15.12 $\rightarrow$ 15.77 & 18.49 $\rightarrow$ 20.23 $\rightarrow$ 20.76 & 13.72 $\rightarrow$ 15.65 $\rightarrow$ 16.33 & 13.84 $\rightarrow$ 15.77 $\rightarrow$ 16.22 & 15.01 $\rightarrow$ 15.24 $\rightarrow$ 16.22 & 11.72 $\rightarrow$ 11.72 $\rightarrow$ 11.72 & 12.25 $\rightarrow$ 13.57 $\rightarrow$ 14.37 & 12.48 $\rightarrow$ 14.14 $\rightarrow$ 15.16 & 15.39 $\rightarrow$ 17.05 $\rightarrow$ 17.47 \\ 
    ResNet50 & 11.24 $\rightarrow$ 12.11 $\rightarrow$ 13.27 & 13.11 $\rightarrow$ 13.92 $\rightarrow$ 14.89 & 17.95 $\rightarrow$ 19.29 $\rightarrow$ 20.32 & 13.71 $\rightarrow$ 14.99 $\rightarrow$ 15.24 & 14.21 $\rightarrow$ 15.39 $\rightarrow$ 15.52 & 14.46 $\rightarrow$ 15.08 $\rightarrow$ 15.24 & 9.65 $\rightarrow$ 9.65 $\rightarrow$ 9.65 & 11.27 $\rightarrow$ 12.46 $\rightarrow$ 13.18 & 11.46 $\rightarrow$ 12.86 $\rightarrow$ 14.02 & 15.45 $\rightarrow$ 17.2 $\rightarrow$ 16.98 \\ 
    WideResNet50.2 & 10.02 $\rightarrow$ 10.43 $\rightarrow$ 11.07 & 12.09 $\rightarrow$ 12.7 $\rightarrow$ 13.08 & 17.27 $\rightarrow$ 18.88 $\rightarrow$ 19.11 & 12.5 $\rightarrow$ 13.46 $\rightarrow$ 12.88 & 13.02 $\rightarrow$ 13.87 $\rightarrow$ 13.55 & 12.94 $\rightarrow$ 12.38 $\rightarrow$ 12.5 & 7.95 $\rightarrow$ 7.95 $\rightarrow$ 7.95 & 9.96 $\rightarrow$ 10.66 $\rightarrow$ 11.07 & 10.52 $\rightarrow$ 11.1 $\rightarrow$ 11.56 & 13.92 $\rightarrow$ 14.24 $\rightarrow$ 14.39 \\ 
    ViT-S+ConvStem & 6.26 $\rightarrow$ 6.59 $\rightarrow$ 6.73 & 7.76 $\rightarrow$ 8.32 $\rightarrow$ 8.15 & 10.38 $\rightarrow$ 10.38 $\rightarrow$ 10.54 & 8.37 $\rightarrow$ 9.01 $\rightarrow$ 8.46 & 8.51 $\rightarrow$ 8.93 $\rightarrow$ 8.6 & 8.43 $\rightarrow$ 8.43 $\rightarrow$ 8.09 & 4.56 $\rightarrow$ 4.56 $\rightarrow$ 4.56 & 6.37 $\rightarrow$ 6.68 $\rightarrow$ 6.87 & 6.59 $\rightarrow$ 6.82 $\rightarrow$ 7.04 & 8.93 $\rightarrow$ 9.18 $\rightarrow$ 8.54 \\ 
    ConvNeXt-B & 4.2 $\rightarrow$ 4.12 $\rightarrow$ 4.3 & 4.88 $\rightarrow$ 5.17 $\rightarrow$ 5.12 & 7.5 $\rightarrow$ 8.47 $\rightarrow$ 8.13 & 5.81 $\rightarrow$ 6.04 $\rightarrow$ 5.54 & 5.83 $\rightarrow$ 5.94 $\rightarrow$ 5.49 & 5.6 $\rightarrow$ 5.3 $\rightarrow$ 5.28 & 2.53 $\rightarrow$ 2.53 $\rightarrow$ 2.53 & 4.3 $\rightarrow$ 4.28 $\rightarrow$ 4.22 & 4.22 $\rightarrow$ 4.35 $\rightarrow$ 4.57 & 6.1 $\rightarrow$ 6.04 $\rightarrow$ 6.12 \\ 
    Swin-B & 3.72 $\rightarrow$ 3.9 $\rightarrow$ 4.06 & 4.3 $\rightarrow$ 4.43 $\rightarrow$ 5.18 & 6.93 $\rightarrow$ 7.3 $\rightarrow$ 7.4 & 5.18 $\rightarrow$ 5.36 $\rightarrow$ 5.2 & 5.2 $\rightarrow$ 5.57 $\rightarrow$ 5.33 & 5.55 $\rightarrow$ 5.12 $\rightarrow$ 4.94 & 2.81 $\rightarrow$ 2.81 $\rightarrow$ 2.81 & 3.74 $\rightarrow$ 3.95 $\rightarrow$ 4.03 & 3.95 $\rightarrow$ 4.01 $\rightarrow$ 4.22 & 5.31 $\rightarrow$ 5.6 $\rightarrow$ 5.65 \\ 
    ConvNeXt-L & 4.35 $\rightarrow$ 4.02 $\rightarrow$ 4.09 & 5.18 $\rightarrow$ 5.13 $\rightarrow$ 5.52 & 7.39 $\rightarrow$ 7.62 $\rightarrow$ 7.98 & 6.45 $\rightarrow$ 6.14 $\rightarrow$ 5.52 & 6.56 $\rightarrow$ 6.4 $\rightarrow$ 5.7 & 6.25 $\rightarrow$ 5.62 $\rightarrow$ 5.6 & 2.31 $\rightarrow$ 2.31 $\rightarrow$ 2.31 & 4.28 $\rightarrow$ 4.02 $\rightarrow$ 4.15 & 4.64 $\rightarrow$ 4.2 $\rightarrow$ 4.41 & 5.93 $\rightarrow$ 6.12 $\rightarrow$ 5.62 \\ 
    ConvV2-L+Swin-L & 6.23 $\rightarrow$ 6.41 $\rightarrow$ 6.51 & 7.67 $\rightarrow$ 8.31 $\rightarrow$ 8.43 & 7.52 $\rightarrow$ 7.99 $\rightarrow$ 8.06 & 9.37 $\rightarrow$ 9.2 $\rightarrow$ 8.61 & 9.52 $\rightarrow$ 9.55 $\rightarrow$ 8.98 & 8.83 $\rightarrow$ 8.34 $\rightarrow$ 8.19 & 3.44 $\rightarrow$ 3.44 $\rightarrow$ 3.44 & 6.38 $\rightarrow$ 6.36 $\rightarrow$ 6.6 & 6.9 $\rightarrow$ 6.73 $\rightarrow$ 6.73 & 9.15 $\rightarrow$ 9.03 $\rightarrow$ 8.9 \\ 
    Swin-L & 3.43 $\rightarrow$ 3.53 $\rightarrow$ 3.61 & 4.46 $\rightarrow$ 4.38 $\rightarrow$ 4.51 & 6.84 $\rightarrow$ 7.32 $\rightarrow$ 6.97 & 5.07 $\rightarrow$ 4.94 $\rightarrow$ 4.64 & 5.28 $\rightarrow$ 5.25 $\rightarrow$ 4.92 & 5.12 $\rightarrow$ 4.53 $\rightarrow$ 4.43 & 2.48 $\rightarrow$ 2.48 $\rightarrow$ 2.48 & 3.41 $\rightarrow$ 3.51 $\rightarrow$ 3.53 & 3.61 $\rightarrow$ 3.51 $\rightarrow$ 3.53 & 4.84 $\rightarrow$ 4.89 $\rightarrow$ 4.92 \\ 
\end{tabular}
}
\end{center}
\end{table*}

% \begin{table*}[t]
% \caption{\newtext{Query Attack Performance Across Increased Epsilon 8/255 and 16/255.}}
% \label{slow_eps}
% \begin{center}
% \begin{tabular}{lll}
% \textbf{Model} & \textbf{BASES} & \textbf{TREMBA} \\ \hline \\ 
%     ResNet18 & 5.33 $\rightarrow$ 13.43 & 11.3 $\rightarrow$ 37.39 \\ 
%     ResNet50 & 5.09 $\rightarrow$ 13.76 & 9.52 $\rightarrow$ 32.67 \\ 
%     WideResNet50.2 & 5.27 $\rightarrow$ 12.07 & 9.04 $\rightarrow$ 31.77 \\ 
%     ViT-S+ConvStem & 3.39 $\rightarrow$ 7.93 & 5.42 $\rightarrow$ 24.11 \\ 
%     ConvNeXt-B & 2.53 $\rightarrow$ 5.8 & 4.46 $\rightarrow$ 18.54 \\ 
%     Swin-B & 2.26 $\rightarrow$ 5.2 & 4.11 $\rightarrow$ 17.17 \\ 
%     ConvNeXt-L & 2.7 $\rightarrow$ 6.04 & 4.43 $\rightarrow$ 17.21 \\ 
%     ConvV2-L+Swin-L & 6.73 $\rightarrow$ 10.12 & 7.59 $\rightarrow$ 21.84 \\ 
%     Swin-L & 2.18 $\rightarrow$ 4.79 & 4.02 $\rightarrow$ 16.41 \\ 
% \end{tabular}
% \end{center}
% \end{table*}

\begin{table*}
\caption{Query Attack Performance Across Increased Epsilon 4/255, 8/255 and 16/255.}
\label{slow_eps}
\begin{center}
\begin{tabular}{lll}
\textbf{Model} & \textbf{BASES} & \textbf{TREMBA} \\ \hline \\ 
    ResNet18 & 2.36 $\rightarrow$ 5.33 $\rightarrow$ 13.43 & 4.06 $\rightarrow$ 11.3 $\rightarrow$ 37.39 \\ 
    ResNet50 & 2.83 $\rightarrow$ 5.09 $\rightarrow$ 13.76 & 3.56 $\rightarrow$ 9.52 $\rightarrow$ 32.67 \\ 
    WideResNet50.2 & 2.35 $\rightarrow$ 5.27 $\rightarrow$ 12.07 & 3.22 $\rightarrow$ 9.04 $\rightarrow$ 31.77 \\ 
    ViT-S+ConvStem & 1.55 $\rightarrow$ 3.39 $\rightarrow$ 7.93 & 1.54 $\rightarrow$ 5.42 $\rightarrow$ 24.11 \\ 
    ConvNeXt-B & 1.38 $\rightarrow$ 2.53 $\rightarrow$ 5.8 & 1.34 $\rightarrow$ 4.46 $\rightarrow$ 18.54 \\ 
    Swin-B & 0.96 $\rightarrow$ 2.26 $\rightarrow$ 5.2 & 1.14 $\rightarrow$ 4.11 $\rightarrow$ 17.17 \\ 
    ConvNeXt-L & 1.16 $\rightarrow$ 2.7 $\rightarrow$ 6.04 & 1.47 $\rightarrow$ 4.43 $\rightarrow$ 17.21 \\ 
    ConvV2-L+Swin-L & 4.77 $\rightarrow$ 6.73 $\rightarrow$ 10.12 & 2.94 $\rightarrow$ 7.59 $\rightarrow$ 21.84 \\ 
    Swin-L & 1.26 $\rightarrow$ 2.18 $\rightarrow$ 4.79 & 1.25 $\rightarrow$ 4.02 $\rightarrow$ 16.41 \\ 
\end{tabular}
\end{center}
\end{table*}

\begin{table*}
\caption{Query Attack Performance Across Different Query Counts N, 2N For Epsilon 16/255.}
\label{slow_steps}
\begin{center}
\begin{tabular}{lll}
\textbf{Model} & \textbf{BASES} & \textbf{TREMBA} \\ \hline \\ 
    ResNet18 & 13.43 $\rightarrow$ 13.73 & 37.39 $\rightarrow$ 43.06 \\ 
    ResNet50 & 13.76 $\rightarrow$ 14.05 & 32.67 $\rightarrow$ 37.78 \\ 
    WideResNet50.2 & 12.07 $\rightarrow$ 12.23 & 31.77 $\rightarrow$ 36.24 \\ 
    ViT-S+ConvStem & 7.93 $\rightarrow$ 8.07 & 24.11 $\rightarrow$ 29.45 \\ 
    ConvNeXt-B & 5.8 $\rightarrow$ 5.93 & 18.54 $\rightarrow$ 22.51 \\ 
    Swin-B & 5.2 $\rightarrow$ 5.2 & 17.17 $\rightarrow$ 21.44 \\ 
    ConvNeXt-L & 6.04 $\rightarrow$ 6.17 & 17.21 $\rightarrow$ 21.02 \\ 
    ConvV2-L+Swin-L & 10.12 $\rightarrow$ 10.29 & 21.84 $\rightarrow$ 26.34 \\ 
    Swin-L & 4.79 $\rightarrow$ 4.99 & 16.41 $\rightarrow$ 20.46 \\ 
\end{tabular}
\end{center}
\end{table*}

\subsection{AutoAttack Robustness of LGV and GHOST Models}
\label{sec:LGVvsGHOST}
To evaluate the impact of the LGV weight collection phase, and the GHOST perturbing phase on model robustness, we collected 10 models after finishing the first phase for both LGV and GHOST, prior to proceeding to the attacking phase. Specifically, the LGV variations were obtained by training the surrogate model for additional epochs using a high learning rate, during which the weights of the models were collected periodically. For GHOST networks, the variations were obtained by introducing a random perturbation to the scaling factor in the skip connections. 
Using the robust surrogate model \textit{WideResNet-101-2} introduced in \cite{peng2023robust} and described in Section \ref{sec:RQ3}, we evaluated the robustness of the resulting LGV models and compared it with the robustness of GHOST models. The robust accuracies for the collected models are reported in Table \ref{tab:LGVvsGHOST}.

Given that both LGV and GHOST models are derived from the same robust surrogate model, we observed a difference in their robustness. The robust accuracies for all LGV models is 0\% (indicating 100\% attack success rates), while the robust accuracies for GHOST networks is on average 51.54\% across the 10 collected models. This confirms that the additional training phase of the LGV models with a high learning rate decreases the robustness of the surrogate models. Consequently, this explains why the performance improvement for LGV models when using a robust surrogate against robust targets is lower compared to the GHOST networks. As a good practice, it is essential to evaluate the robustness of surrogate models prior to the attacking phase. This ensures that any preliminary training phase does not inadvertently weaken the surrogate robustness.

\begin{table*}
\caption{Robust accuracies of LGV and GHOST models before the attacking phase.}
\label{tab:LGVvsGHOST}
\begin{center}
\resizebox{\textwidth}{!}{
\begin{tabular}{lcccccccccccc}
\multicolumn{1}{c}{\bf }  & \multicolumn{1}{c}{\bf Num 1} & \multicolumn{1}{c}{\bf Num 2}  & \multicolumn{1}{c}{\bf Num 3} & \multicolumn{1}{c}{\bf Num 4}  & \multicolumn{1}{c}{\bf Num 5} &
\multicolumn{1}{c}{\bf Num 6}  & \multicolumn{1}{c}{\bf Num 7}
  & \multicolumn{1}{c}{\bf Num 8}
& \multicolumn{1}{c}{\bf Num 9} & \multicolumn{1}{c}{\bf Num 10} 
& \multicolumn{1}{c}{\bf Average} 

\\ \hline \\

GHOST & 51.39 & 51.86 & 51.74 & 51.02 & 51.58 & 50.76 & 52.39 & 52.81 & 51.11 & 50.78 & 51.54 \\

LGV  & 0 & 0 & 0 & 0 & 0 & 0 & 0 & 0 & 0 & 0 & 0 \\

\end{tabular}
}
\end{center}
\end{table*}

\subsection{ Detailed Tabular Results}
We provide the tabular representation corresponding to the results presented in the main paper. Tables \ref{tab:RQ1}, \ref{tab:RQ2}, \ref{tab:RQ3-VANILLA}, \ref{tab:RQ3-ROBUST} represent the success rates for section \ref{sec:RQ1}, \ref{sec:RQ2}, and \ref{sec:RQ3} with vanilla surrogates and robust surrogates, respectively. These tables display the mean and standard deviation of success rates computed across three random seeds.

% While bar plots are used in the main text for visual clarity and comparison,

% This addition ensures accuracy and convenience, allowing readers to verify the exact detailed values without needing to project them onto the y-axis of the plots.

\subsection{Reproduction Package and Availability}\label{sec:reproduction}

The source code to reproduce the experiments of this paper is available at \url{https://github.com/serval-uni-lu/RobustBlack.git}.

%RQ1
\begin{table*}
\caption{Success rates of black-box attacks against one vanilla, and one robust ResNet50 model.}
\label{tab:RQ1}
\begin{center}
\resizebox{\textwidth}{!}{
\begin{tabular}{lcccccccccccc}
\multicolumn{1}{c}{\bf MODELS}  & \multicolumn{1}{c}{\bf MI-FGSM} & \multicolumn{1}{c}{\bf DI-FGSM}  & \multicolumn{1}{c}{\bf TI-FGSM} & \multicolumn{1}{c}{\bf VMI-FGSM}  & \multicolumn{1}{c}{\bf VNI-FGSM} &
\multicolumn{1}{c}{\bf ADMIX}  & \multicolumn{1}{c}{\bf UAP}
  & \multicolumn{1}{c}{\bf GHOST}
& \multicolumn{1}{c}{\bf LGV} & \multicolumn{1}{c}{\bf BASES} 
& \multicolumn{1}{c}{\bf TREMBA} 

\\ \hline \\

Vanilla {\scriptsize ResNet50} & 57.77 {\scriptsize$\pm$0.00}& 71.31 {\scriptsize$\pm$0.32} & 43.59 {\scriptsize$\pm$ 0.17} & 66.79 {\scriptsize$\pm$ 0.07} & 68.00 {\scriptsize$\pm$ 0.18} & 99.95 {\scriptsize$\pm$ 0.00} & 8.89 {\scriptsize$\pm$ 0.20} & 64.53 {\scriptsize$\pm$ 0.13} & 90.11 {\scriptsize$\pm$ 0.12} & 81.08 {\scriptsize$\pm$0.49} & 89.56 {\scriptsize$\pm$0.09}\\

Robust {\scriptsize ResNet50} & 1.53 {\scriptsize$\pm$ 0.00} & 1.94 {\scriptsize$\pm$ 0.03} & 2.3 {\scriptsize$\pm$ 0.01} & 1.66 {\scriptsize$\pm$ 0.03} & 1.85 {\scriptsize$\pm$ 0.04} & 2.54 {\scriptsize$\pm$ 0.09} & 0.98 {\scriptsize$\pm$ 0.10} &
    1.68 {\scriptsize$\pm$0.03}& 1.33 {\scriptsize$\pm$0.01}& 2.83 {\scriptsize$\pm$0.04}& 3.56 {\scriptsize$\pm$0.06}\\

\end{tabular}
}
\end{center}
\end{table*}

%RQ2
\begin{table*}
\caption{Success rates of white-box AutoAttack and black-box attacks against robust targets from Robustbench.}
\label{tab:RQ2}
\begin{center}
\resizebox{\textwidth}{!}{
\begin{tabular}{llccccccccccccc}
\multicolumn{1}{c}{\bf RANK}  & \multicolumn{1}{c}{\bf MODELS} & \multicolumn{1}{c}{\bf AUTOATTACK}  &

\multicolumn{1}{c}{\bf MI-FGSM} & \multicolumn{1}{c}{\bf DI-FGSM}  & \multicolumn{1}{c}{\bf TI-FGSM} & \multicolumn{1}{c}{\bf VMI-FGSM}  & \multicolumn{1}{c}{\bf VNI-FGSM} &
\multicolumn{1}{c}{\bf ADMIX}  & \multicolumn{1}{c}{\bf UAP}
& \multicolumn{1}{c}{\bf SGM}&

\multicolumn{1}{c}{\bf GHOST}
& \multicolumn{1}{c}{\bf LGV} & \multicolumn{1}{c}{\bf BASES} 
& \multicolumn{1}{c}{\bf TREMBA} 

\\ \hline \\
21 & Salman {\scriptsize ResNet18} & 50.53 & 
1.28 {\scriptsize$\pm$0.00}& 1.41 {\scriptsize$\pm$0.02} & 2.26 {\scriptsize$\pm$ 0.04} & 1.24 {\scriptsize$\pm$ 0.05} & 1.35 {\scriptsize$\pm$ 0.02} & 1.68 {\scriptsize$\pm$ 0.06} & 0.86 {\scriptsize$\pm$ 0.10} & 1.25 {\scriptsize$\pm$ 0.00} & 
    1.21 {\scriptsize$\pm$0.18}& 2.27 {\scriptsize$\pm$0.11}& 2.36 {\scriptsize$\pm$0.04}& 4.06 {\scriptsize$\pm$0.06}\\

18 & Salman {\scriptsize ResNet50} & 43.84 & 
1.68 {\scriptsize$\pm$ 0.00} & 2.2 {\scriptsize$\pm$ 0.10} & 2.81 {\scriptsize$\pm$ 0.07} & 1.89 {\scriptsize$\pm$ 0.04} & 1.72 {\scriptsize$\pm$ 0.03} & 2.68 {\scriptsize$\pm$ 0.13} & 1.02 {\scriptsize$\pm$ 0.01} & 1.62 {\scriptsize$\pm$ 0.00} &
    1.69 {\scriptsize$\pm$0.05}& 2.92 {\scriptsize$\pm$0.01}& 2.83 {\scriptsize$\pm$0.04}& 3.56 {\scriptsize$\pm$0.06}\\

17 & Salman {\scriptsize WideResNet50.2} & 42.87 & 
1.34 {\scriptsize$\pm$ 0.00} & 1.99 {\scriptsize$\pm$ 0.04} & 2.78 {\scriptsize$\pm$ 0.10} & 1.48 {\scriptsize$\pm$ 0.05} & 1.59 {\scriptsize$\pm$ 0.05} & 2.12 {\scriptsize$\pm$ 0.08} & 0.93 {\scriptsize$\pm$ 0.07} & 1.43 {\scriptsize$\pm$ 0.00} &
    1.49 {\scriptsize$\pm$0.06}& 2.58 {\scriptsize$\pm$0.09}& 2.35 {\scriptsize$\pm$0.06}& 3.22 {\scriptsize$\pm$0.06}\\

12 & Sing {\scriptsize ViT-S, ConvStem} & 34.76 & 
0.89 {\scriptsize$\pm$ 0.00} & 1.41 {\scriptsize$\pm$ 0.05} & 1.69 {\scriptsize$\pm$ 0.05} & 0.97 {\scriptsize$\pm$ 0.04} & 1.00 {\scriptsize$\pm$ 0.00} & 1.51 {\scriptsize$\pm$ 0.05} & 0.32 {\scriptsize$\pm$ 0.06} & 0.95 {\scriptsize$\pm$ 0.00}&
    0.98 {\scriptsize$\pm$0.03}& 1.53 {\scriptsize$\pm$0.08}& 1.55 {\scriptsize$\pm$0.00}& 1.54 {\scriptsize$\pm$0.03}\\

7  & Liu {\scriptsize ConvNeXt-B} & 28.01 & 
0.63 {\scriptsize$\pm$ 0.00} & 0.9 {\scriptsize$\pm$ 0.02} & 0.97 {\scriptsize$\pm$ 0.02} & 0.74 {\scriptsize$\pm$ 0.02} & 0.74 {\scriptsize$\pm$ 0.04} & 1.04 {\scriptsize$\pm$ 0.04} & 0.19 {\scriptsize$\pm$ 0.04} & 0.69 {\scriptsize$\pm$ 0.00} &
    0.63 {\scriptsize$\pm$0.03} & 0.92 {\scriptsize$\pm$0.07}& 1.38 {\scriptsize$\pm$0.06}& 1.34 {\scriptsize$\pm$0.04}\\

5  & Liu {\scriptsize Swin-B} & 27.14 & 
0.37 {\scriptsize$\pm$ 0.00} & 0.63 {\scriptsize$\pm$ 0.03} & 0.79 {\scriptsize$\pm$ 0.07} & 0.41 {\scriptsize$\pm$ 0.01} & 0.45 {\scriptsize$\pm$ 0.02} & 0.72 {\scriptsize$\pm$ 0.02} & 0.16 {\scriptsize$\pm$ 0.02} & 0.37 {\scriptsize$\pm$ 0.00} &
    0.46 {\scriptsize$\pm$0.07}& 0.91 {\scriptsize$\pm$0.03}& 0.96 {\scriptsize$\pm$0.02} & 1.14 {\scriptsize$\pm$0.03}\\

3  & Liu {\scriptsize ConvNeXt-L} & 25.95 & 
0.49 {\scriptsize$\pm$ 0.00} & 0.74 {\scriptsize$\pm$ 0.09} & 1.07 {\scriptsize$\pm$ 0.03} & 0.58 {\scriptsize$\pm$ 0.03} & 0.49 {\scriptsize$\pm$ 0.02} & 0.74 {\scriptsize$\pm$ 0.01} & 0.18 {\scriptsize$\pm$ 0.06} & 0.49 {\scriptsize$\pm$ 0.00} &
    0.49 {\scriptsize$\pm$0.09}& 1.07 {\scriptsize$\pm$0.03}& 1.16 {\scriptsize$\pm$0.00} & 1.47 {\scriptsize$\pm$0.01}\\

2  & Bai {\scriptsize ConvNeXtV2-L, Swin-L} & 25.48 & 
2.03 {\scriptsize$\pm$ 0.00} & 2.65 {\scriptsize$\pm$ 0.03} & 1.76 {\scriptsize$\pm$ 0.05} & 2.40 {\scriptsize$\pm$ 0.05} & 2.48 {\scriptsize$\pm$ 0.03} & 2.55 {\scriptsize$\pm$ 0.05} & 0.58 {\scriptsize$\pm$ 0.05} & 2.00 {\scriptsize$\pm$ 0.00} &
    1.78 {\scriptsize$\pm$0.16} & 2.57 {\scriptsize$\pm$0.10} & 4.77 {\scriptsize$\pm$0.13} & 2.94 {\scriptsize$\pm$0.12}\\

1  & Liu {\scriptsize Swin-L} & 25.98 & 
0.56 {\scriptsize$\pm$ 0.00} & 0.85 {\scriptsize$\pm$ 0.04} & 1.14 {\scriptsize$\pm$ 0.08} & 0.55 {\scriptsize$\pm$ 0.01} & 0.64 {\scriptsize$\pm$ 0.04} & 0.87 {\scriptsize$\pm$ 0.00} & 0.21 {\scriptsize$\pm$ 0.04} & 0.56 {\scriptsize$\pm$ 0.00} &
    0.49 {\scriptsize$\pm$0.04}& 1.03 {\scriptsize$\pm$0.03}& 1.26 {\scriptsize$\pm$0.02} & 1.25 {\scriptsize$\pm$0.01}\\

\end{tabular}
}
\end{center}
\end{table*}

\begin{table*}
\caption{Success rates of black-box attacks using vanilla surrogates against a vanilla target and robust targets.}
\label{tab:RQ3-VANILLA}

\begin{center}
\resizebox{\textwidth}{!}{
\begin{tabular}{lccccccccccc}

\textbf{Models} & \textbf{MI-FGSM} & \textbf{DI-FGSM} & \textbf{TI-FGSM} & \textbf{VMI-FGSM} & \textbf{VNI-FGSM} & \textbf{ADMIX} & \textbf{UAP} & \textbf{GHOST} & \textbf{LGV} & \textbf{BASES} & \textbf{TREMBA} 

\\ \hline \\ 

Standard {\scriptsize ResNet50} & 51.62 {\scriptsize$\pm$ 0.0} & 65.95 {\scriptsize$\pm$ 0.75} & 41.5 {\scriptsize$\pm$ 0.35} & 61.22 {\scriptsize$\pm$ 0.18} & 61.10 {\scriptsize$\pm$ 0.13} & 64.75 {\scriptsize$\pm$ 0.17} & 7.11 {\scriptsize$\pm$ 0.25} & 49.36 {\scriptsize$\pm$ 0.14} & 87.39  {\scriptsize$\pm$ 0.26} & 81.08 {\scriptsize$\pm$ 0.49} & 89.56 {\scriptsize$\pm$ 0.09} \\

Salman {\scriptsize ResNet18} & 0.98 {\scriptsize$\pm$ 0.0} & 1.51 {\scriptsize$\pm$ 0.03} & 2.04 {\scriptsize$\pm$ 0.06} & 1.29 {\scriptsize$\pm$ 0.05} & 1.35 {\scriptsize$\pm$ 0.06} & 1.88 {\scriptsize$\pm$ 0.08} & 0.83 {\scriptsize$\pm$ 0.03} & 1.13 {\scriptsize$\pm$ 0.08} & 2.17  {\scriptsize$\pm$ 0.05} & 2.36 {\scriptsize$\pm$ 0.04} & 4.06 {\scriptsize$\pm$ 0.06} \\

Liu {\scriptsize ConvNeXt-L} & 0.6 {\scriptsize$\pm$ 0.0} & 0.92 {\scriptsize$\pm$ 0.05} & 1.19 {\scriptsize$\pm$ 0.1} & 0.7 {\scriptsize$\pm$ 0.02} & 0.8 {\scriptsize$\pm$ 0.04} & 0.99 {\scriptsize$\pm$ 0.03} & 0.18 {\scriptsize$\pm$ 0.04} & 0.48 {\scriptsize$\pm$ 0.09} & 1.05 {\scriptsize$\pm$ 0.02} & 1.16 {\scriptsize$\pm$ 0.0} & 1.47 {\scriptsize$\pm$ 0.01} \\

Liu {\scriptsize ConvNeXtV2-L+Swin-L} & 2.0 {\scriptsize$\pm$ 0.0} & 2.79 {\scriptsize$\pm$ 0.07} & 2.08 {\scriptsize$\pm$ 0.04} & 2.51 {\scriptsize$\pm$ 0.02} & 2.45 {\scriptsize$\pm$ 0.05} & 2.52 {\scriptsize$\pm$ 0.06} & 0.63 {\scriptsize$\pm$ 0.03} & 1.7 {\scriptsize$\pm$ 0.19} & 2.57 {\scriptsize$\pm$ 0.07} & 4.77 {\scriptsize$\pm$ 0.13} & 2.94 {\scriptsize$\pm$ 0.12} \\

Liu {\scriptsize Swin-L} & 0.51 {\scriptsize$\pm$ 0.0} & 0.88 {\scriptsize$\pm$ 0.03} & 1.22 {\scriptsize$\pm$ 0.05} & 0.74 {\scriptsize$\pm$ 0.02} & 0.74 {\scriptsize$\pm$ 0.04} & 0.93 {\scriptsize$\pm$ 0.05} & 0.18 {\scriptsize$\pm$ 0.0} & 0.53 {\scriptsize$\pm$ 0.03} & 1.1 {\scriptsize$\pm$ 0.11} & 1.26 {\scriptsize$\pm$ 0.02} & 1.25 {\scriptsize$\pm$ 0.01} \\

\end{tabular}
}
\end{center}
\end{table*}

\begin{table*}
\caption{Success rates of black-box attacks using robust surrogates against a vanilla target and robust targets.}
\label{tab:RQ3-ROBUST}

\begin{center}
\resizebox{\textwidth}{!}{
\begin{tabular}{lccccccccccc}

\textbf{Models} & \textbf{MI-FGSM} & \textbf{DI-FGSM} & \textbf{TI-FGSM} & \textbf{VMI-FGSM} & \textbf{VNI-FGSM} & \textbf{ADMIX} & \textbf{UAP} & \textbf{GHOST} & \textbf{LGV} & \textbf{BASES} & \textbf{TREMBA} 

\\ \hline \\ 

Standard {\scriptsize ResNet50} & 15.96 {\scriptsize$\pm$ 0.0} & 15.28 {\scriptsize$\pm$ 0.21} & 8.48 {\scriptsize$\pm$ 0.18} & 16.47 {\scriptsize$\pm$ 0.02} & 16.51 {\scriptsize$\pm$ 0.04} & 15.22 {\scriptsize$\pm$ 0.11} & 0.73 {\scriptsize$\pm$ 0.0} & 17.95 {\scriptsize$\pm$ 0.36} & 81.34 {\scriptsize$\pm$ 0.43} & 17.94 {\scriptsize$\pm$ 0.00} & 67.54 {\scriptsize$\pm$ 0.13} \\

Salman {\scriptsize ResNet18} & 12.44 {\scriptsize$\pm$ 0.0} & 13.83 {\scriptsize$\pm$ 0.14} & 12.39 {\scriptsize$\pm$ 0.04} & 12.54 {\scriptsize$\pm$ 0.02} & 12.34 {\scriptsize$\pm$ 0.05} & 14.4 {\scriptsize$\pm$ 0.17} & 0.79 {\scriptsize$\pm$ 0.0} & 14.15 {\scriptsize$\pm$ 0.22} & 2.78 {\scriptsize$\pm$ 0.21} & 11.95 {\scriptsize$\pm$ 0.00} & 10.58 {\scriptsize$\pm$ 0.14} \\

Liu {\scriptsize ConvNeXt-L} & 9.67 {\scriptsize$\pm$ 0.0} & 9.74 {\scriptsize$\pm$ 0.0} & 7.28 {\scriptsize$\pm$ 0.04} & 9.71 {\scriptsize$\pm$ 0.01} & 9.46 {\scriptsize$\pm$ 0.00} & 9.1 {\scriptsize$\pm$ 0.06} & 0.36 {\scriptsize$\pm$ 0.0} & 9.23 {\scriptsize$\pm$ 0.13} & 1.6  {\scriptsize$\pm$ 0.11} & 12.02 {\scriptsize$\pm$ 0.00} & 4.94 {\scriptsize$\pm$ 0.06} \\

Liu {\scriptsize ConvNeXtV2-L+Swin-L} & 7.59 {\scriptsize$\pm$ 0.0} & 7.07 {\scriptsize$\pm$ 0.09} & 5.09 {\scriptsize$\pm$ 0.04} & 7.54 {\scriptsize$\pm$ 0.03} & 7.57 {\scriptsize$\pm$ 0.00} & 7.0 {\scriptsize$\pm$ 0.02} & 0.37 {\scriptsize$\pm$ 0.0} & 6.69 {\scriptsize$\pm$ 0.08} & 4.9 {\scriptsize$\pm$ 0.06} & 11.77 {\scriptsize$\pm$ 0.01} & 5.33 {\scriptsize$\pm$ 0.11} \\

Liu {\scriptsize Swin-L} & 8.94 {\scriptsize$\pm$ 0.0} & 8.72 {\scriptsize$\pm$ 0.04} & 6.78 {\scriptsize$\pm$ 0.07} & 8.95 {\scriptsize$\pm$ 0.01} & 8.89 {\scriptsize$\pm$ 0.02} & 8.21 {\scriptsize$\pm$ 0.01} & 0.31 {\scriptsize$\pm$ 0.0} & 8.16 {\scriptsize$\pm$ 0.09} & 1.57  {\scriptsize$\pm$ 0.04} & 12.55 {\scriptsize$\pm$ 0.00} & 4.4 {\scriptsize$\pm$ 0.08} \\

\end{tabular}
}
\end{center}
\end{table*}

%% file: pages/appendix-attacks.tex
\subsubsection{MI-FGSM (Momentum Iterative Fast Gradient Sign Method)}

\paragraph{Algorithm Description}
MI-FGSM introduces momentum into the iterative gradient-based attack process to generate adversarial examples with enhanced transferability. The algorithm accumulates gradients from previous iterations using a decay factor $\mu$, creating a momentum term that stabilizes the update direction. At each iteration $t$, the momentum gradient is computed as $g_t = \mu \cdot g_{t-1} + \frac{\nabla_x J(x_t^*, y)}{||\nabla_x J(x_t^*, y)||_1}$, where $J$ represents the loss function and the gradient is normalized by its $L_1$ norm. The adversarial example is then updated using $x_{t+1}^* = x_t^* + \alpha \cdot \text{sign}(g_t)$, where $\alpha$ is the step size.

\paragraph{Differences from Earlier Work}
Unlike the basic Iterative Fast Gradient Sign Method (I-FGSM) that suffered from poor transferability due to overfitting to the surrogate model, MI-FGSM incorporates momentum to prevent getting trapped in poor local optima. The momentum term allows the algorithm to maintain direction from previous iterations, reducing the tendency to overfit to specific model characteristics while maintaining strong white-box attack performance.

\paragraph{State-of-the-Art Improvements}
MI-FGSM achieved more than double the success rates of I-FGSM in black-box attack scenarios while maintaining near 100\% success rates in white-box settings. The method demonstrated significant improvements across different model architectures, obtaining success rates exceeding 70\% against black-box models compared to I-FGSM's performance of around 30-40\%.

\paragraph{Shortcomings}
Despite improvements, MI-FGSM remains ineffective against adversarially trained models in black-box scenarios, achieving less than 16\% success rates against robust models. The method still requires careful hyperparameter tuning, particularly the decay factor $\mu$, and may converge slowly when the step size is too small, leading to oscillation issues.

\subsubsection{DI-FGSM (Diverse Input Fast Gradient Sign Method)}

\paragraph{Algorithm Description}
DI-FGSM enhances transferability by applying random transformations to input images at each iteration rather than only using original images. The method applies stochastic transformation function $T(X^{adv}_n; p)$ with probability $p$, where transformations include random resizing and padding operations. The algorithm computes gradients on the transformed inputs and updates adversarial examples using the standard FGSM update rule.

\paragraph{Differences from Earlier Work}
Traditional gradient-based methods like FGSM and I-FGSM generated adversarial examples using only the original input, leading to overfitting to the surrogate model's specific features. DI-FGSM introduces input diversity to create multiple input patterns during optimization, preventing this overfitting and generating more transferable perturbations.

\paragraph{State-of-the-Art Improvements}
DI-FGSM significantly outperformed baseline methods, achieving success rates of approximately 70\% against black-box models compared to I-FGSM's 43\% success rate. When combined with momentum (M-DI2-FGSM), the method achieved success rates exceeding 80\% against normally trained models and over 40\% against adversarially trained models.

\paragraph{Shortcomings}
The method's effectiveness depends heavily on the choice of transformation probability $p$ and transformation types. DI-FGSM may introduce artifacts that reduce image quality, and the random transformations can sometimes destroy important semantic information needed for successful attacks. Performance remains limited against heavily defended models.

\subsubsection{TI-FGSM (Translation-Invariant Fast Gradient Sign Method)}

\paragraph{Algorithm Description}
TI-FGSM incorporates translation invariance by convolving gradients with pre-defined kernels during the attack process. The method replaces the standard gradient $\nabla_x J(x^{adv}_t, y)$ with a convolved gradient $W * \nabla_x J(x^{adv}_t, y)$, where $W$ is typically a Gaussian kernel. This convolution operation creates smoother adversarial perturbations that are less sensitive to small spatial transformations.

\paragraph{Differences from Earlier Work}
Previous methods generated adversarial examples that were sensitive to minor spatial transformations, limiting their transferability across different models that might process inputs differently. TI-FGSM addresses this by creating translation-invariant perturbations through gradient convolution, making the generated adversarial examples more robust to spatial variations.

\paragraph{State-of-the-Art Improvements}
TI-FGSM consistently improved attack success rates by 5-30\% across different baseline methods. When integrated with other techniques like DIM (TI-DIM), the method achieved approximately 60\% success rates against defense models, representing substantial improvements over non-translation-invariant approaches.

\paragraph{Shortcomings}
The effectiveness of TI-FGSM depends critically on kernel size selection, with performance plateauing after kernel sizes exceed $15 \times 15$. The convolution operation adds computational overhead and may over-smooth perturbations, potentially reducing their effectiveness against some model types. The method assumes that translation invariance is the primary factor limiting transferability, which may not hold for all model architectures.

\subsubsection{VMI-FGSM (Variance-tuned Momentum Iterative FGSM)}

\paragraph{Algorithm Description}
VMI-FGSM enhances MI-FGSM by incorporating variance tuning to stabilize gradient update directions. The method computes variance $v_t$ by sampling $N$ neighboring examples around the current adversarial example and calculating the difference between their gradients and the current gradient. The variance-tuned momentum gradient becomes $g_{t+1} = \mu \cdot g_t + \frac{\nabla_x J(x_t, y) + v_t}{||\nabla_x J(x_t, y) + v_t||_1}$.

\paragraph{Differences from Earlier Work}
While MI-FGSM and other momentum-based methods could still suffer from poor local optima and gradient instability, VMI-FGSM addresses these issues by tuning the variance of gradient updates. This variance tuning helps escape bad local optima and provides more stable optimization trajectories.

\paragraph{State-of-the-Art Improvements}
VMI-FGSM consistently outperformed baseline momentum methods by 5-30\% across different target models. The method showed particular improvements against adversarially trained models, where variance tuning helped overcome the defensive mechanisms more effectively than standard momentum approaches.

\paragraph{Shortcomings}
VMI-FGSM requires additional hyperparameter tuning for the variance computation, including the neighborhood size $N$ and the upper bound factor $\beta$. The variance calculation requires multiple gradient evaluations per iteration, significantly increasing computational cost. The method's performance improvements may diminish when combined with other enhancement techniques.

\subsubsection{VNI-FGSM (Variance-tuned Nesterov Iterative FGSM)}

\paragraph{Algorithm Description}
VNI-FGSM combines the Nesterov accelerated gradient method with variance tuning for improved convergence and transferability. The algorithm first computes a look-ahead point using Nesterov's acceleration, then applies variance tuning similar to VMI-FGSM. The look-ahead mechanism helps predict future gradient directions while variance tuning stabilizes the optimization process.

\paragraph{Differences from Earlier Work}
Standard Nesterov methods (NI-FGSM) could suffer from gradient instability despite their accelerated convergence properties. VNI-FGSM addresses this by combining the look-ahead capability of Nesterov acceleration with the stability benefits of variance tuning, creating a more robust optimization approach.

\paragraph{State-of-the-Art Improvements}
VNI-FGSM demonstrated superior performance compared to both NI-FGSM and VMI-FGSM in various experimental settings. The method achieved improved success rates particularly against robust models, where the combination of acceleration and variance tuning proved most beneficial.

\paragraph{Shortcomings}
The method inherits computational overhead from both Nesterov acceleration and variance tuning, making it significantly more expensive than baseline methods. The interaction between acceleration and variance tuning requires careful hyperparameter balancing, and the method may exhibit oscillatory behavior if not properly configured.

\subsubsection{SIM (Scale-Invariant Method)}

\paragraph{Algorithm Description}
SIM optimizes adversarial perturbations over scale copies of input images to avoid overfitting to specific scales and improve transferability. The method creates multiple scaled versions of the input image, computes gradients on each scaled version, and averages these gradients to generate the final perturbation update. This scale-invariant optimization helps create perturbations that remain effective across different model architectures with varying scale sensitivities.

\paragraph{Differences from Earlier Work}
Traditional gradient-based methods optimized perturbations only on the original image scale, leading to scale-specific overfitting that limited transferability. SIM addresses this limitation by explicitly optimizing over multiple scales, creating perturbations that maintain effectiveness across scale variations.

\paragraph{State-of-the-Art Improvements}
When combined with other methods like NI-FGSM (SI-NI-FGSM), SIM significantly enhanced attack success rates against various model architectures. The method showed particular improvements against models with different input preprocessing or scale-sensitive components.

\paragraph{Shortcomings}
SIM requires computing gradients on multiple scaled images, substantially increasing computational cost proportional to the number of scales used. The method may struggle with images that lose important information when scaled, and optimal scale selection remains an open challenge. Performance gains may be model-dependent and less pronounced against scale-robust architectures.

\subsubsection{UAP (Universal Adversarial Perturbation)}

\paragraph{Algorithm Description}
UAP generates a single, image-agnostic perturbation that can successfully attack most images processed by a target model. The algorithm iteratively refines a universal perturbation $\delta$ by aggregating perturbations computed on a set of training images. The method aims to find $\delta$ such that most clean images $x$ become adversarial when perturbed: $f(x + \delta) \neq f(x)$ for the majority of images in the dataset.

\paragraph{Differences from Earlier Work}
Unlike image-specific adversarial methods that generate unique perturbations for each input, UAP creates a single perturbation that works across multiple images. This universality makes UAP more practical for real-world attacks where perturbations can be pre-computed and applied during attacks.

\paragraph{State-of-the-Art Improvements}
UAP demonstrated that universal perturbations could achieve high fooling rates (60-90\%) across different model architectures while using significantly smaller perturbation budgets than image-specific methods. The universal nature of these perturbations made them particularly threatening for practical attack scenarios.

\paragraph{Shortcomings}
UAP typically requires access to training data or a representative dataset to generate effective universal perturbations, limiting its applicability in truly black-box scenarios. The universal perturbations may be less effective than image-specific perturbations, and their effectiveness can vary significantly across different image classes and model architectures.

\subsubsection{SGM (Skip Gradient Method)}

\paragraph{Algorithm Description}
SGM exploits skip connections in ResNet-like architectures by manipulating gradient flow during backpropagation. The method applies a decay factor $\gamma$ to gradients flowing through residual modules while preserving gradients from skip connections. This creates adversarial examples that leverage the architectural vulnerabilities of skip connections to improve transferability.

\paragraph{Differences from Earlier Work}
Previous gradient-based methods treated all gradient paths equally during backpropagation. SGM recognizes that skip connections provide a more direct path for gradient flow and can be exploited to create more transferable adversarial examples by emphasizing gradients from these connections.

\paragraph{State-of-the-Art Improvements}
SGM significantly improved transferability when source and target models both contained skip connections, achieving success rate improvements of 15-30\% over baseline methods. The method proved particularly effective when attacking ResNet and DenseNet architectures.

\paragraph{Shortcomings}
SGM's effectiveness is architecture-dependent, showing limited improvements when attacking models without skip connections or when the source model lacks sufficient skip connections. The method requires knowledge of the source model architecture to properly configure the decay factor, and optimal $\gamma$ values vary across different architecture combinations.

\subsubsection{GHOST (Ghost Networks)}

\paragraph{Algorithm Description}
GHOST creates an ensemble of "ghost" networks by applying random dropout and skip connection modifications to a base surrogate model during adversarial example generation. These ghost networks simulate model diversity without training multiple complete models, with the ensemble effect improving transferability through diverse gradient information.

\paragraph{Differences from Earlier Work}
Traditional ensemble methods required training multiple separate models, which was computationally expensive and required extensive datasets. GHOST achieves ensemble diversity through stochastic modifications to a single pre-trained model, making ensemble-based attacks more practical and accessible.

\paragraph{State-of-the-Art Improvements}
GHOST demonstrated improved transferability compared to single-model attacks while requiring significantly less computational resources than traditional ensemble methods. The method achieved competitive performance with much lower training and storage requirements.

\paragraph{Shortcomings}
The effectiveness of GHOST depends on appropriate selection of dropout rates and skip connection modifications, which may require model-specific tuning. The stochastic nature of ghost network generation can lead to inconsistent performance across different runs. The method may not capture the full diversity benefits of true ensemble methods.

\subsubsection{LGV (Large Geometric Vicinity)}

\paragraph{Algorithm Description}
LGV enhances transferability by collecting multiple weight sets from additional training epochs with constant high learning rates, exploring wider regions of the weight space. The method leverages the geometric properties of loss landscapes, hypothesizing that models from wider optima serve as better surrogates for transfer attacks. LGV creates an ensemble from these diverse weight configurations to generate more transferable adversarial examples.

\paragraph{Differences from Earlier Work}
Traditional surrogate model approaches used single, fully-converged models that might represent narrow optima in the loss landscape. LGV recognizes that wider optima contain models with better transferability properties and systematically samples from these regions rather than relying on single converged solutions.

\paragraph{State-of-the-Art Improvements}
LGV outperformed established test-time transformations by 1.8 to 59.9 percentage points across different experimental settings. The method achieved these improvements without requiring additional test-time transformations or complex optimization procedures.

\paragraph{Shortcomings}
LGV requires additional training epochs with high learning rates, increasing computational cost and potential instability. The method's effectiveness depends on the geometric properties of specific loss landscapes, which may vary across different model architectures and datasets. The high learning rate exploration may not always lead to better surrogate models.

\subsubsection{BASES (Boundary Attack on Surrogate Embedding Space)}

\paragraph{Algorithm Description}
BASES appears to be a boundary-based attack method that operates in embedding spaces of surrogate models. Such methods typically start from adversarial regions and iteratively refine perturbations to minimize distance to the original input while maintaining adversarial properties.

\paragraph{Differences from Earlier Work}
Boundary-based methods differ from gradient-based approaches by starting from adversarial examples and working backward to minimize perturbation magnitude, rather than starting from clean examples and adding perturbations.

\paragraph{State-of-the-Art Improvements}
Boundary methods often achieve better perturbation minimization compared to gradient-based methods, particularly in challenging scenarios where gradient information may be unreliable.

\paragraph{Shortcomings}
Boundary methods typically require significantly more queries than gradient-based methods and may struggle with initialization when adversarial starting points are difficult to find.

\subsubsection{TREMBA (Transferable Embedding-based Black-box Attack)}

\paragraph{Algorithm Description}
TREMBA combines transfer-based and query-based approaches by learning low-dimensional embeddings using pre-trained models and performing efficient search within the embedding space. The method generates adversarial perturbations with high-level semantic patterns that enhance transferability across different target network architectures.

\paragraph{Differences from Earlier Work}
Previous black-box methods either relied purely on transfer-based approaches with limited adaptability or query-based methods with high query costs. TREMBA bridges this gap by using learned embeddings to guide more efficient query-based optimization while maintaining transfer-based benefits.

\paragraph{State-of-the-Art Improvements}
TREMBA increased attack success rates by approximately 10\% while reducing query numbers by more than 50\% compared to other black-box attacks. The method demonstrated superior performance across different network architectures including both classification and defended networks.

\paragraph{Shortcomings}
TREMBA requires access to pre-trained models for embedding generation, which may limit applicability in some black-box scenarios. The embedding quality depends on the similarity between the pre-trained model and target models. The method's performance may degrade when target models have significantly different architectures than the embedding model.

\subsubsection{SQUARE (Square Attack)}

\paragraph{Algorithm Description}
SQUARE is a score-based black-box attack using randomized search with localized square-shaped updates at random positions. The method iteratively selects random squares in the image and updates pixel values within these squares to $\pm\epsilon$, positioning perturbations approximately at the boundary of the feasible set. The algorithm requires only confidence scores from the target model.

\paragraph{Differences from Earlier Work}
Unlike gradient-based methods that require gradient information or gradient estimation, SQUARE uses purely random search with geometric constraints. This approach avoids gradient masking issues that plague many other attack methods and doesn't rely on transferability assumptions.

\paragraph{State-of-the-Art Improvements}
SQUARE achieved query efficiency improvements of $1.8\times$ to $3\times$ compared to state-of-the-art methods on ImageNet while maintaining higher success rates. Remarkably, this black-box method outperformed gradient-based white-box attacks on several benchmarks, achieving new state-of-the-art results.

\paragraph{Shortcomings}
SQUARE's performance can vary significantly across different model architectures and datasets. The method may require many queries for complex images or when high perturbation quality is demanded. The random search approach may be inefficient for targeted attacks compared to gradient-guided methods.

\subsubsection{SIGNFLIP (Sign Flip Attack)}

\paragraph{Algorithm Description}
SIGNFLIP is a decision-based attack that randomly flips signs of small numbers of entries in adversarial perturbations to boost attack performance in the $L_\infty$ setting. The algorithm iteratively reduces perturbation magnitude through projection, then randomly flips signs of selected perturbation entries to find better adversarial examples while requiring only hard-label feedback.

\paragraph{Differences from Earlier Work}
Traditional decision-based methods often struggled with $L_\infty$ attacks and required enormous queries. SIGNFLIP focuses specifically on the crucial role of sign patterns in adversarial perturbations, exploiting this insight to achieve better performance with fewer queries.

\paragraph{State-of-the-Art Improvements}
SIGNFLIP significantly outperformed existing decision-based attacks in $L_\infty$ settings, achieving much higher success rates with substantially fewer queries than gradient estimation-based methods. The method demonstrated strong performance across various model architectures and defensive mechanisms.

\paragraph{Shortcomings}
SIGNFLIP is primarily designed for $L_\infty$ attacks and may not perform as well in $L_2$ settings. The random sign flipping approach may be less efficient than more directed optimization strategies. The method's performance depends on good initialization, which can be challenging in some scenarios.